\pdfoutput=1

\documentclass[11pt]{article}

\usepackage{emnlp2022}

\usepackage{times}
\usepackage{latexsym}

\usepackage[T1]{fontenc}

\usepackage[utf8]{inputenc}

\usepackage{microtype}

\usepackage{subcaption}
\usepackage{graphicx}
\usepackage{graphics}
\usepackage{tabularx}
\usepackage{multirow}
\usepackage{booktabs}
\usepackage{amsmath}
\usepackage{bbold}
\usepackage{bm}
\usepackage{relsize}
\usepackage[para]{threeparttable}

\newcommand{\cocorrespondingfootnote}{
    \let\oldthefootnote=\thefootnote
    \setcounter{footnote}{1}
    \renewcommand{\thefootnote}{\#}
    \footnotetext{Co-corresponding authors.}
    \let\thefootnote=\oldthefootnote
}

\newcommand{\astfootnote}[1]{
    \let\oldthefootnote=\thefootnote
    \setcounter{footnote}{1}
    \renewcommand{\thefootnote}{\fnsymbol{footnote}}
    \footnotetext{#1}
    \let\thefootnote=\oldthefootnote
}

\title{Ground-Truth Labels Matter: \\ A Deeper Look into Input-Label Demonstrations}

\author{Kang Min Yoo$^{*\#\dagger\ddagger\mathsection}$, Junyeob Kim$^{*\mathsection}$, Hyuhng Joon Kim$^\mathsection$, Hyunsoo Cho$^\mathsection$, \\
\textbf{Hwiyeol Jo$^\ddagger$, Sang-Woo Lee$^{\dagger\ddagger\natural}$, Sang-goo Lee$^\mathsection$, Taeuk Kim$^{\#\mathparagraph}$}\\
 $^\mathsection$Seoul National University,
 $^\dagger$NAVER AI Lab, $^\ddagger$NAVER CLOVA \\
  $^\natural$Korea Advanced Institute of Science and Technology, 
 $^\mathparagraph$Hanyang University \\
 \texttt{\{juny116,heyjoonkim,johyunsoo,sglee\}@europa.snu.ac.kr}\\
 \texttt{\{hwiyeol.jo,sang.woo.lee,kangmin.yoo\}@navercorp.com}\\
 \texttt{kimtaeuk@hanyang.ac.kr} \\ 
 }
 
\begin{document}
\maketitle
\begin{abstract}
Despite recent explosion of interests in in-context learning, the underlying mechanism and the precise impact of the quality of demonstrations remain elusive.
Intuitively, ground-truth labels should have as much impact in in-context learning (ICL) as supervised learning, but recent work reported that the input-label correspondence is significantly less important than previously thought.
Intrigued by this counter-intuitive observation, we re-examine the importance of ground-truth labels in in-context learning.
With the introduction of two novel metrics, namely Label-Correctness Sensitivity and Ground-truth Label Effect Ratio (GLER), we were able to conduct quantifiable analysis on the impact of ground-truth label demonstrations.
Through extensive analyses, we find that the correct input-label mappings can have varying impacts on the downstream in-context learning performances, depending on the experimental configuration.
Through additional studies, we identify key components, such as the verbosity of prompt templates and the language model size, as the controlling factor to achieve more noise-resilient ICL.

\end{abstract}

\astfootnote{Equal contributions.}
\cocorrespondingfootnote

\section{Introduction}

Large-scale language models \cite{rae2021scaling,chowdhery2022palm,smith2022using,thoppilan2022lamda} have shaped the NLP scene by introducing in-context learning (ICL) \cite{brown2020language} as a novel approach to adapt language models for downstream tasks without explicit fine-tuning. ICL enables language models to learn and predict from task-specific prompts that contain demonstrations in the natural language format, despite the language models were only trained to predict the next word token. Inspired by the new discovery, a flurry of recent work has investigated ways to explain and exploit the ICL mechanism (\citet{schick2020exploiting,lu2021fantastically}; \textit{inter alia}), but it remains elusive.

\begin{figure}[t]
    \centering
    \includegraphics[width=0.50\textwidth]{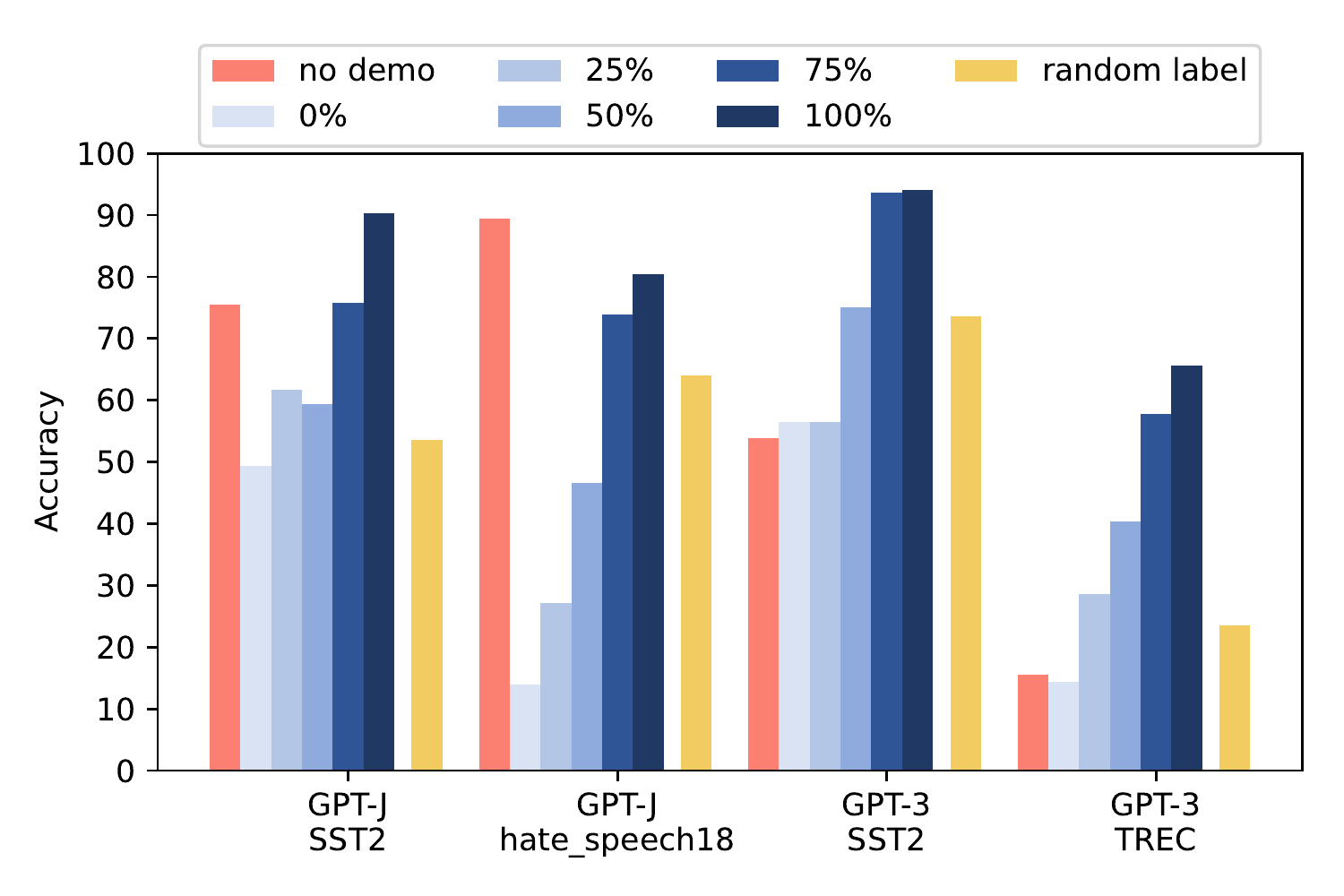}
    \caption{A demonstration of cases where the effect of the ground-truth label in in-context learning is much more significant than the aggregated results reported by \citet{min2022rethinking}.}
    \label{fig:main1}
\end{figure}

\citet{min2022rethinking} have recently re-evaluated the role of input-label correspondence in demonstrations for ICL.
Specifically, the authors have shown that the correct mapping between input and its label contributes less to the final performance than we thought compared to other aspects, including the format of demonstrations and the awareness of the input and label space.
This finding is intriguing and has been sensational, as it is counter-intuitive to the expectation of how statistical learning typically works in supervised settings, and therefore it shows a potential of exploiting (few-shot) in-context learning given no real training data.
For example, prior work established the strong impact of example ordering \citep{zhao2021calibrate}, hence in-context learning being less sensitive to the correctness of label demonstrations, which forms the basis of supervised learning, seems contradictory.

However, we encountered cases where the observation is inconsistent with the recent finding on the matter (Figure \ref{fig:main1}).
Specifically, we found that the difference between the performance from the ground-truth label demonstration and that from entirely incorrect labels was as large as 80\% (accuracy) for the hate speech dataset \cite{de2018hate} on GPT-J \cite{gpt-j}.
Similar observations were found with the larger GPT-3 \cite{brown2020language} model and other datasets (TREC \cite{li-roth-2002-learning}).
These cases illustrate how sensitive in-context learning can be to label demonstrations depending on the ICL settings.
Thus, we cast a doubt on whether the trend can be generalized in diverse configurations, raising a call for an in-depth analysis of the phenomenon.

In this paper, we revisit the findings of \citet{min2022rethinking} and take a closer look into the importance of ground-truth labels for in-context learning.
First, we point out limitations of the existing work.
Then, we introduce novel metrics, namely Label-Correctness Sensitivity and Ground-Truth Label Effect Ratio (GLER), to reveal that the input-label correspondence plays a more vital role in contextual demonstration than previously considered.
Furthermore, we show that the trend contradictory to the previous discovery becomes salient if we diverge the experimental settings (e.g., datasets, metrics, and templates) from the previous work.
We observe the same trend in various language models, such as GPT-J and GPT-3 \cite{brown2020language}.

In addition, this paper uses statistics to provide a systematic and complementary perspective to the existing findings on the label-demonstration impact.
To be specific, we combine linear regression and auxiliary metrics to conduct all-around and deeper analyses on how the ICL classification performance changes against label-demonstration corruption.
To do so, we define the notion of sensitivity to quantify the degree to which the downstream classification performance changes when a model is subject to a fixed amount of label corruption.
As a result, we demonstrate several noticeable patterns that support the claim that there is a considerable relationship between the performance and label correctness.
It is worth noting that this trend was not clearly visible in the previous work, where the results of each dataset are macro-averaged rather than individually analyzed.

However, insensitivity, or robustness, towards the incorrectness of label-demonstrations is a useful property to have for many situations. For example, when augmenting an extremely small number of (e.g., less than four) examples using data augmentation techniques, exhibiting performance resilience towards prompt templates that consist of noisy synthetic examples as demonstrations is desirable. We further analyze how different factors of ICL, such as the inference method, the underlying language model, and the adoption of advanced ICL strategies, affect the performance sensitivity towards noises in input-label demonstrations, paving the way for a new approach to exploiting the demonstration insensitivity.

In summary, our contributions are as follows.

\begin{itemize}
    \item We re-examine the recent findings on the phenomenon that the ICL performance is insensitive towards input-label demonstrations.
    \item We propose two new quantifiable metrics, sensitivity and GLER, to measure the impact of ground-truth label demonstrations on ICL.
    \item We conduct a thorough examination of how different components of ICL could impact the model's insensitivity towards label noises, allowing future work to exploit such property.
\end{itemize}

\section{Looking Deeper into Ground-Truth Labels}
\label{sec:looking}

\begin{figure*}[t]
    \centering
    \begin{subfigure}{0.45\textwidth}
        \centering
        \includegraphics[width=\textwidth]{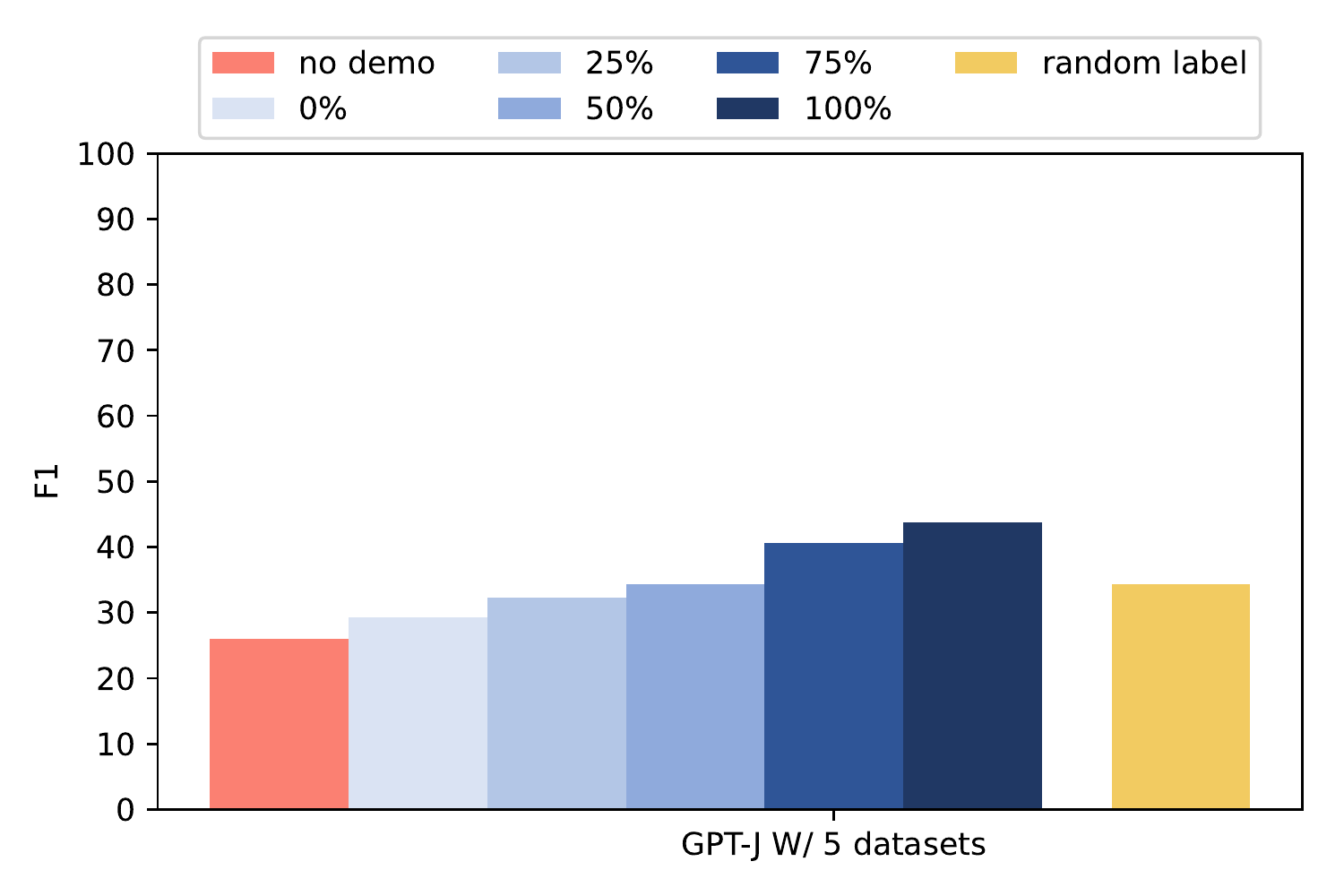}
    \end{subfigure}
    \hfill
    \begin{subfigure}{0.45\textwidth}
        \centering
        \includegraphics[width=\textwidth]{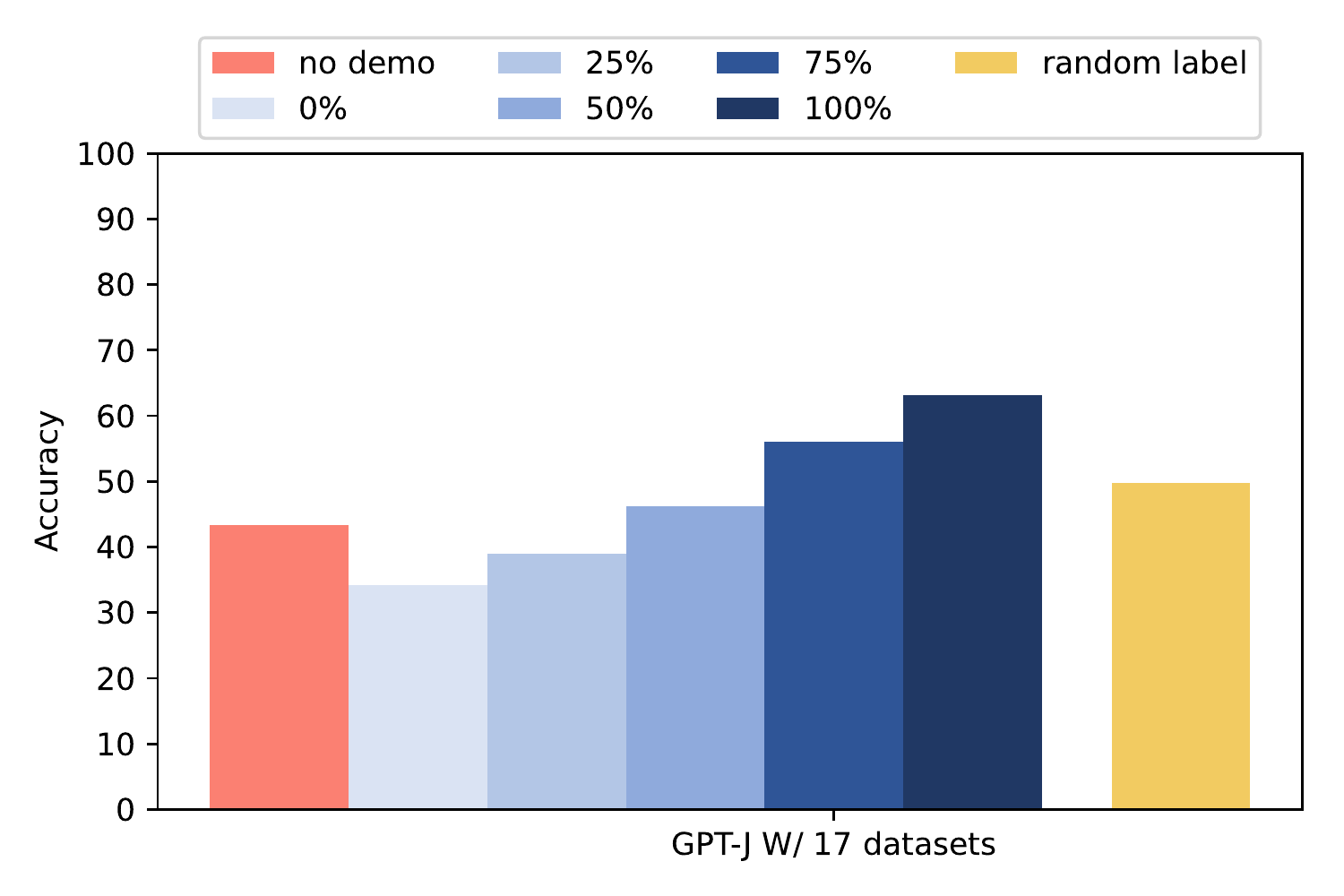}
    \end{subfigure}
    \caption{A counter-example of slightly varied but equally valid experimental settings is shown on the right, while the results from the prior experimental settings \cite{min2022rethinking} is shown on the left. ``No Demo'' refers to the result without demonstrations and ``Random Label'' refers to the result with label demonstrations replaced with a random label uniformly sampled from the label space. Minor variations in the experimental settings could result in a large difference in the degree of which the ICL performance responds to the label corruption. More details on the experiment is described in Appendix \ref{app:counter-example}.}
    \label{fig:counter-example}
\end{figure*}

Demonstrations of ground-truth labels\footnote{Here, \textit{label demonstrations} refer to the demonstration of input-label correspondence and not the demonstration of label space.}, correctly paired with inputs, have been known to be a crucial factor of supervised learning, but a recent work by \citet{min2022rethinking} purportedly revealed the possibly counter-intuitive nature of label demonstrations in in-context learning (ICL). Specifically, the findings implied that the correctness of input-label correspondence in in-context demonstrations is not as important as we have thought. We name this phenomenon \textit{input-label insensitivity}. Although the finding was supported by reasonably large-scale experiments, covering various experimental variables such as datasets, language models, in-context learning types, etc., we found that, through \textit{deeper analysis} of the experiments, input-label insensitivity is not consistent across all experimental settings.

This section highlights the limitations of the existing work, proposes new metrics to quantify the impact of input-label correspondence, and finally presents deeper analyses of the ICL experiments utilizing the newly proposed metrics.

\subsection{Limitations of the Existing Work}

\label{sec:looking-limitations}

\citet{min2022rethinking} showed that replacing ground-truth labels in prompt demonstrations with incorrect labels marginally affects the \textit{mean-aggregated overall} performance on selected datasets. Although the input-label insensitivity phenomenon was less prominent on GPT-J with the direct ICL method, the ICL still performed better when entirely incorrect labels were given than the absence of demonstrations (the zero-shot baseline), allegedly supporting the input-label sensitivity idea \citep{min2022rethinking}. However, we argue that there are mainly two limitations to the existing claim. 

\paragraph{Over-generalization} The existing claim suffers from over-generalization in two regards: (1) the mean-aggregated results fails to capture the insensitivity behavior in individual tasks and (2) the proposed experimental settings in the existing work is not general enough to be fully supportive of the claim. Mean-aggregation does not paint the full picture without the information on the variance. 
Furthermore, individual analyses on large-scale tasks are needed to obtain precise insight into input-label sensitivity. 
Our deeper analyses on the ICL experiments (\S \ref{sec:looking-analyses}) provide more evidence of this claim.

The second over-generalization is supported by the existence of a counter-example: higher input-label sensitivity observed from a slight varied but equally valid experimental settings (Figure \ref{fig:counter-example}). The subfigure on the left corresponds to the result of an existing set of experimental settings, where the Noisy Channel method \cite{min2021noisy} was used for ICL, the macro-F1 score for the evaluation metric, and the five classification datasets listed in the existing work. The subfigure on the right has been obtained using (\textit{Direct}) method, the accuracy score as metric and results were aggregated from all 17 datasets listed in the existing work (see Appendix\ref{app:counter-example}). 

\paragraph{Lack of Quantification} Existing work relies on human judgement to determine the input-label sensitivity, which could be subjective. Furthermore, we are not only interested in \textit{whether} the input-label insensitivity phenomenon exists but also \textit{how} insensitive the ICL is towards the demonstrations, enabling us to exploit the phenomenon. Hence, a set of systematic quantification methods is needed to perform the deeper analyses.

\subsection{Key Concepts}

This subsection establishes key concepts and notations related to our analysis on the impact of input-label demonstrations and the downstream ICL performance. $x$ and $c$ denote the input and the label respectively. They exist in each respective input ($\mathcal{X}$) or label space ($\mathcal{C}$) associated with the dataset or task. A language model $P$ predicts the next token given the preceding tokens: $P (x_t | x_{<t})$. In ICL, a prompt $\mathcal{P}$ is designed to elicit particular behaviors from the language model. For example, to utilize the language model as a text classifier, a prompt template $\mathcal{T}$ takes a set of examples $\mathcal{D}_\text{ex} = \{ (x_1, c_1), ..., (x_k, c_k) \}$ and a test input $x$ to produce the prompt $\mathcal{P}$. The prompt is then fed into the language model to produce the most plausible continuation: $\text{argmax}_{x'} P(x' | \mathcal{P})$. A task-specific verbalizer $\mathcal{V}$ is designed to interpret the generated output $x'$ into the label space $\mathcal{C}$. We measure the performance $y$ of the language model $P$ and the prompt template $\mathcal{T}$ on a test set $\mathcal{D}_\text{test}$.

Our analyses mainly involve manipulating $\mathcal{T}$ and the example set $\mathcal{D}_\text{ex}$ to set-up baselines and conduct ablation studies. Key experimental set-ups include: \textbf{No Demo}, or denoted as ``zero-shot'', represents zero-shot predictions, where the prompt template $\mathcal{T}$ ignores $\mathcal{D}_\text{ex}$ and only uses the test input $x$: $P(c|x)$. The example set $\mathcal{D}_\text{ex}$ in \textbf{$\alpha$\%-Correct} consists $k \times a / 100$ correct input-label pairs and $k \times (1 - a/100)$ incorrect pairs where $(0 \leq a \leq 100)$. For \textbf{Random Label}, the labels $c$ in $\mathcal{D}_\text{ex}$ are replaced by uniform samples from the label space $\mathcal{C}$, and it is one of the key baselines of our studies. Additional details on the set-up variations are presented in Appendix \ref{app:counter-example}.

\subsection{Metrics for Measuring the Impact of Input-Label Demonstrations}

\label{sec:looking-metrics}

This section proposes two new metrics to quantify the impact of input-label demonstrations in ICL.

\paragraph{Label-Correctness Sensitivity} We define label-correctness sensitivity, or \textbf{sensitivity} for short, as the degree of which the downstream classification performance changes when the model is subject to a fixed amount of label corruption. Sensitivity in the context of in-context learning demonstrations can be computed by conducting the single-scalar linear regression analysis on a performance metric (e.g., accuracy or F1-score) $y$ against the percentage of examples that are labelled correctly ($s$):

$$
y=\beta_0 +\beta_1 s
$$

\noindent where $\beta_0$ is the bias and $\beta_1$ is the coefficient of label correctness. The scalar value of the weight parameter $\beta_1$ is interpreted as the sensitivity measure. The data points for linear regression were obtained by following the experimental protocol proposed by \citet{min2022rethinking}. The sensitivity measure can be interpreted as a linearly interpolated measure of performance degradation for each unit decrease in label correctness.

\paragraph{Ground-Truth Label Effect Ratio (GLER)} Another way to understand the impact of labels, namely correct or ground-truth labels, is to quantify \textit{how much the ground-truth labels improve the ICL performance compared to the random-label baseline}. The higher the gap, the bigger the impact the ground-truth labels have on the performance. The gap is then normalized by the performance difference between ground-truth labels and the absence-of-demonstration baseline (zero-shot):

\begin{equation} \label{eq:gler}
\text{GLER} = \frac{y_\text{GT} - y_\text{RL}}{y_\text{GT} - y_\emptyset}
\end{equation}

\noindent where $y_\text{GT}$ is the ground-truth label performance, $y_\text{RL}$ the random-label baseline (\textbf{Random-Label}), and $y_\emptyset$ the zero-shot performance. The denominator in Equation \ref{eq:gler} is intended to allow the GLER metric to be compared across different tasks. Additionally, we clip GLER to be bounded between 0 and 1.

\subsection{Deeper Analyses}

\label{sec:looking-analyses}

This subsection performs deeper analyses using the aforementioned metrics to reveal additional insights into input-label insensitivity.

\subsubsection{Experimental Setup}

All of our experiments mentioned in the rest of the paper generally follows the experimental settings in \citet{min2022rethinking}, where \textbf{$\alpha$\%-Correct} is mainly utilized to conduct sensitivity analysis. However, there are key differences: (1) we do not employ label-length normalization (in our experiments length normalization does not always increase the performance), and there are minor template $\mathcal{T}$ design differences, including how the separator token interacts with the model and the dataset-specific implementation of data preprocessor; (2) we use accuracy, instead of F1-score, as the primary evaluation metrics for ICL performance. However, we do report the full results in Appendix \ref{app:counter-example}, along with the full details of the setup.

\subsubsection{Label Correctness Does Affect Performance}

\begin{table}[t]
    \centering
    \resizebox{0.99\columnwidth}{!}{%
    \begin{tabular}{lccc}
        \toprule
        Method & Coefficient & Intercept & $R^2$ \\
        \midrule
        GPT-NeoX Direct & 0.300 & 0.327 & 0.810  \\
        GPT-J Direct & 0.309  & 0.291  & 0.861 \\
        \bottomrule
    \end{tabular}
    }
    \caption{Aggregated linear regression analysis on the performance against the percentage of correct labels. ``Ours'' indicates that the data points for the linear regression analysis were obtained using our proposed experimental settings (Appendix \ref{app:counter-example}).}
    \label{tab:linreg}
\end{table}

To analyze the overall sensitivity of performance under the variation of label correctness, we aggregate sensitivities across all 17 classification datasets and the results are shown in Table \ref{tab:linreg}. The results show that the aggregated sensitivity is significantly high with good fit (in the range of 0.81-0.86) for all configurations. When tested on our specific setup, the sensitivity was as high as 0.309, implying that, on average, there was a 0.309\% drop in accuracy for each percentage drop in label correctness.

The trend of sensitivity, which is more apparent in our quantitative analysis, may have been overlooked due to the relative dwarfing effect from zero-shot (or ``no demo") results in prior studies. The results also show that the sensitivity is lower in the Channel method,\footnote{We hypothesize that this observation is attributed to the fact that, while generating longer sentences, prediction distribution from Channel model are more affected by the pre-trained prior rather than the current context.} suggesting that sensitivity can be significantly lowered with the employment of more advanced ICL methods.

\subsubsection{Label Demonstration Impact is Highly Varied Across Tasks and Settings}

\begin{figure}[t]
    \centering
    \begin{subfigure}{0.45\textwidth}
        \centering
        \begin{subfigure}{\textwidth}
            \centering
            \includegraphics[width=\textwidth]{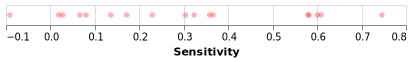}
        \end{subfigure}
        \hfill
        \begin{subfigure}{\textwidth}
            \centering
            \includegraphics[width=\textwidth]{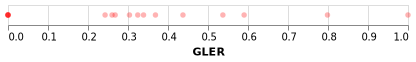}
        \end{subfigure}
        \hfill
        \begin{subfigure}{\textwidth}
            \centering
            \includegraphics[width=\textwidth]{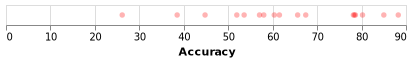}
        \end{subfigure}
        \caption{GPT-NeoX}
    \end{subfigure}
    \centering
    \begin{subfigure}{0.45\textwidth}
        \centering
        \begin{subfigure}{\textwidth}
            \centering
            \includegraphics[width=\textwidth]{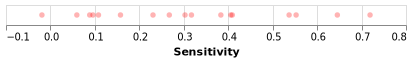}
        \end{subfigure}
        \hfill
        \begin{subfigure}{\textwidth}
            \centering
            \includegraphics[width=\textwidth]{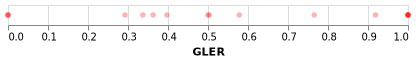}
        \end{subfigure}
        \hfill
        \begin{subfigure}{\textwidth}
            \centering
            \includegraphics[width=\textwidth]{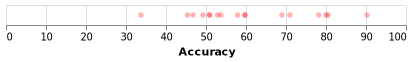}
        \end{subfigure}
        \caption{GPT-J}
    \end{subfigure}
    \caption{Individual scatter-plots of the proposed metrics, sensitivity and GLER, across two models (GPT-NeoX and GPT-J) and 17 datasets. We also report the nominal ground-truth label accuracy values to further showcase the highly varied nature of the tasks.}
    \label{fig:sensitivity}
\end{figure}

Although the aggregating analysis shows a general sensitive trend towards demonstration correctness, individual analyses shed deeper insight into the distribution of task sensitivities.
Individual sensitivity plots are illustrated in Figure \ref{fig:sensitivity}. Sensitivity can vary from small negative values (indicating increasing performance under increasing label corruption) to value as high as 0.815 (for the hate speech dataset), suggesting that summarizing the trend for all tasks and datasets may be difficult and that certain datasets may possess distributional properties that allow models to more easily exploit label demonstrations. This high-variance observation is valid for other metrics (GLER and the ground-truth label performance) as well. Further analyses are available in \S\ref{sec:when}

\subsubsection{Sensitivity and Task Difficulty}
\label{sec:looking-analyses-sensitivity}

\begin{figure}[t]
    \centering
    \includegraphics[width=0.4\textwidth]{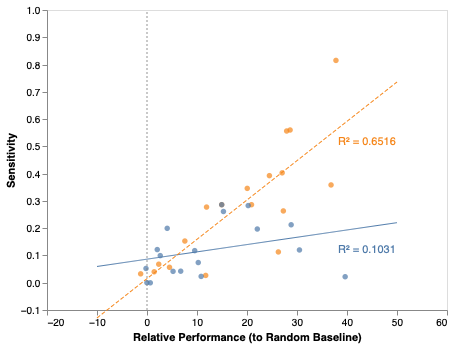}
    \caption{A scatter plot of sensitivities of 17 datasets against the corresponding task difficulties measured using the relative performance. The Direct approach is colored in orange and the Channel approach is colored in blue. The dashed vertical line indicates a neutral performance level where there is no difference with the random baselines. More details is found in Appendix \ref{app:task-difficulty}.}
    \label{fig:sensitivity-difficulty}
\end{figure}

Tasks where the model struggle to exploit in-context demonstrations may exhibit low sensitivity towards them, since understanding patterns in demonstrations is inherently linked with the ability to absorb demonstrative label-supervision. To confirm our theory, we conduct an analysis on the sensitivities of 17 datasets against the task difficulty. We define task difficulty as the relative performance of ground-truth label demonstrations compared to a baseline. Specifically, relative performance $y_\text{rel}$ is computed by $y_\text{rel}=y_\text{GT} - y_\text{baseline}$. We consider the \textit{random baseline}.

Our analysis (Figure \ref{fig:sensitivity-difficulty}) shows that the model’s performance sensitivity is strongly related to the difficulty of the task. The tasks, where the model exhibits low sensitivity (i.e. $<0.1$), struggle to achieve meaningful classification performance. This suggests that designing experiments with datasets that can be meaningfully solved using in-context learners may be more important than previously understood. Hence, the sensitivity measure by itself is insufficient for benchmarking the impact of input-label demonstrations.

\section{When Do the Ground-Truth Labels Actually (Not) Matter?}
\label{sec:when}

As revealed in our deeper analyses (\S \ref{sec:looking-analyses}), many factors including datasets and the choice of the ICL method can significantly affect the label-sensitivity. Gaining more understanding of the mechanism by which the input-label correspondence impacts the downstream ICL performance could enable us to systematically exploit the label-insensitivity phenomenon. For example, few-shot ICL models can be improved to tolerate label noises from synthetic data samples generated in the joint input and label spaces \cite{yoo2021gpt3mix}.

To understand the conditions that reduce the label sensitivity, we conduct a series of experiments that investigate different factors contribute to the phenomenon quantified using the metrics proposed in \S \ref{sec:looking-metrics}. Namely, we consider the particular technical choice in carrying out ICL (whether to employ the noisy channel method \citep{min2021noisy} and the likelihood calibration \citep{zhao2021calibrate}), various properties of the prompt template (the number of in-context examples and the verbosity), and the model size.

\paragraph{Sensitivity and GLER} Recall that the sensitivity measure is the nominal coefficient of the linear line fitted on the performance-versus-label-corruption data points. Since baselines can vary depending on the experimental setting, hyperparameters and the dataset\footnote{For example, under the same conditions (GPT-J and Direct inference), the random-label accuracy baseline is 28.08 for TREC and 53.58 for SST2.}, comparing the nominal sensitivity alone can be inconclusive, as the same degree of absolute improvement has different implications depending on the baseline level. To account for the variations in the \textit{characteristics} of the task and the model, we consider GLER and the ground-truth label performance as the auxiliary measures in the following studies.

\subsection{Techniques for In-context Learning}

\label{sec:when-icl}

\begin{figure}
    \centering
    \begin{subfigure}{0.45\textwidth}
        \centering
        \begin{subfigure}{\textwidth}
            \centering
            \includegraphics[width=\textwidth]{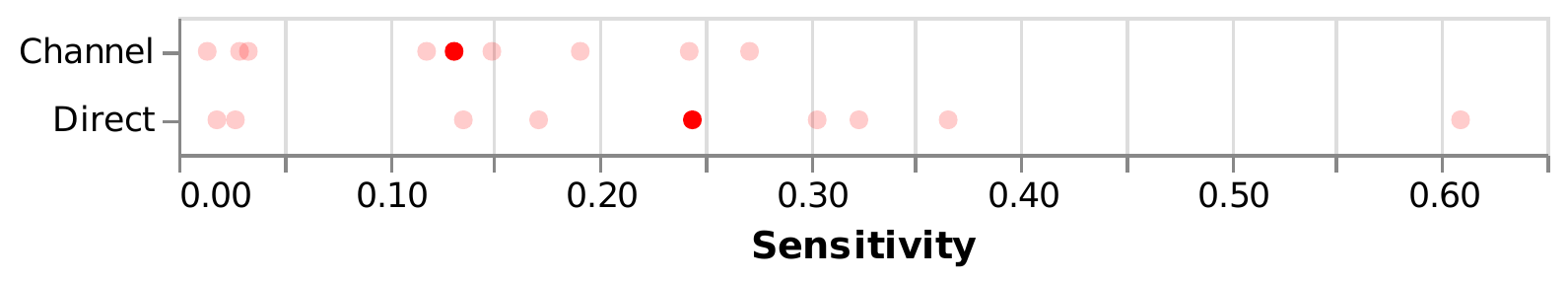}
        \end{subfigure}
        \hfill
        \begin{subfigure}{\textwidth}
            \centering
            \includegraphics[width=\textwidth]{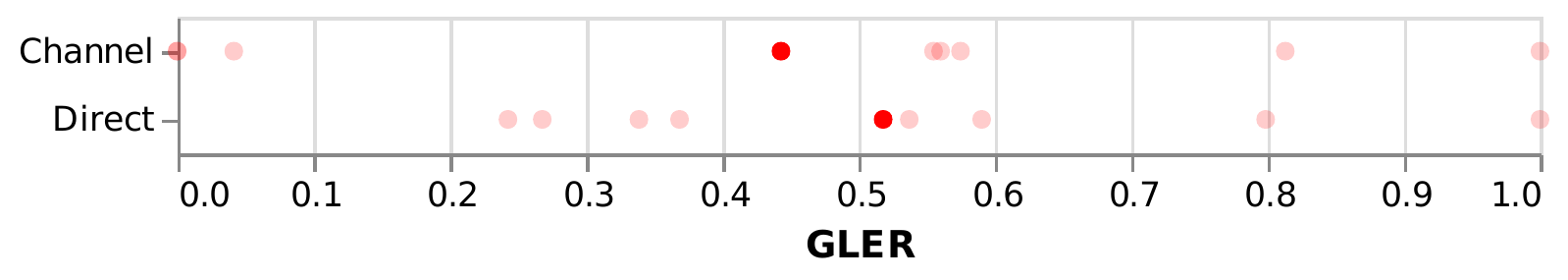}
        \end{subfigure}
        \hfill
        \begin{subfigure}{\textwidth}
            \centering
            \includegraphics[width=\textwidth]{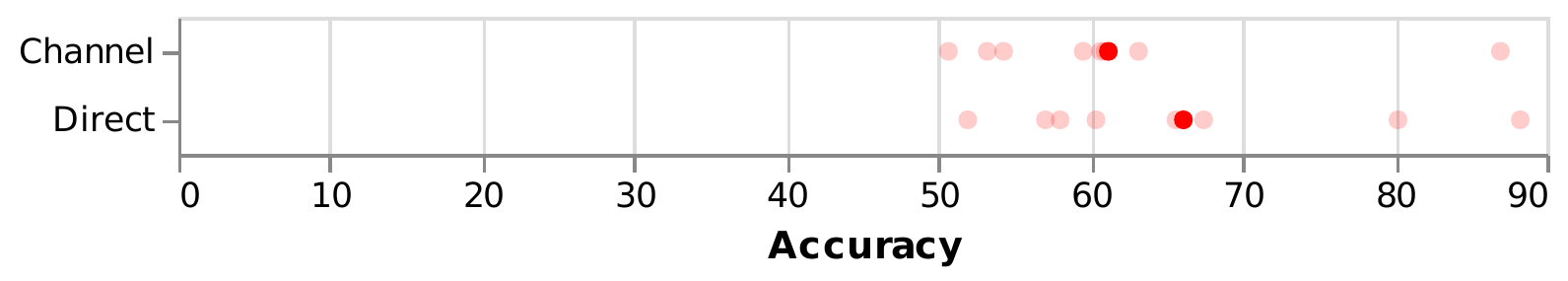}
        \end{subfigure}
        \caption{GPT-NeoX}
    \end{subfigure}
    \centering
    \begin{subfigure}{0.45\textwidth}
        \centering
        \begin{subfigure}{\textwidth}
            \centering
            \includegraphics[width=\textwidth]{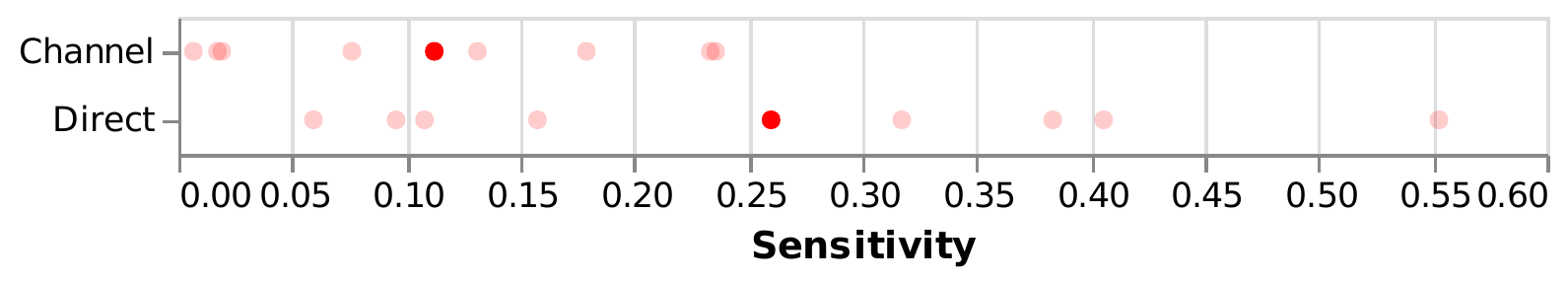}
        \end{subfigure}
        \hfill
        \begin{subfigure}{\textwidth}
            \centering
            \includegraphics[width=\textwidth]{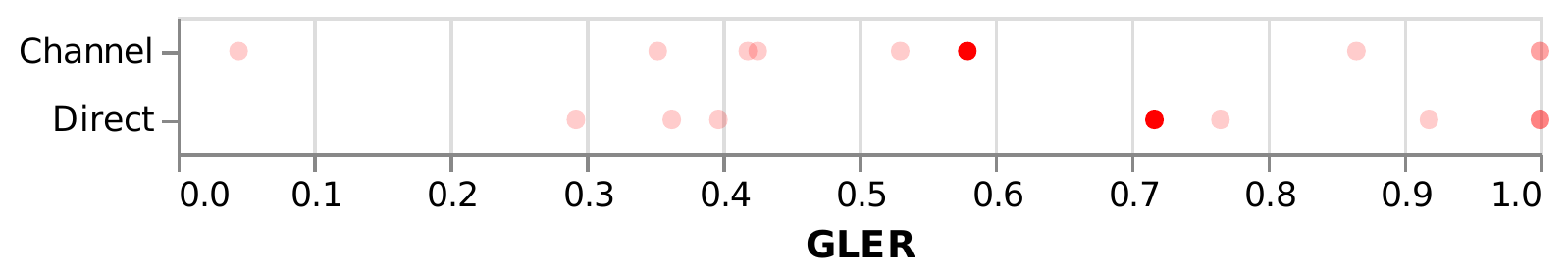}
        \end{subfigure}
        \begin{subfigure}{\textwidth}
            \centering
            \includegraphics[width=\textwidth]{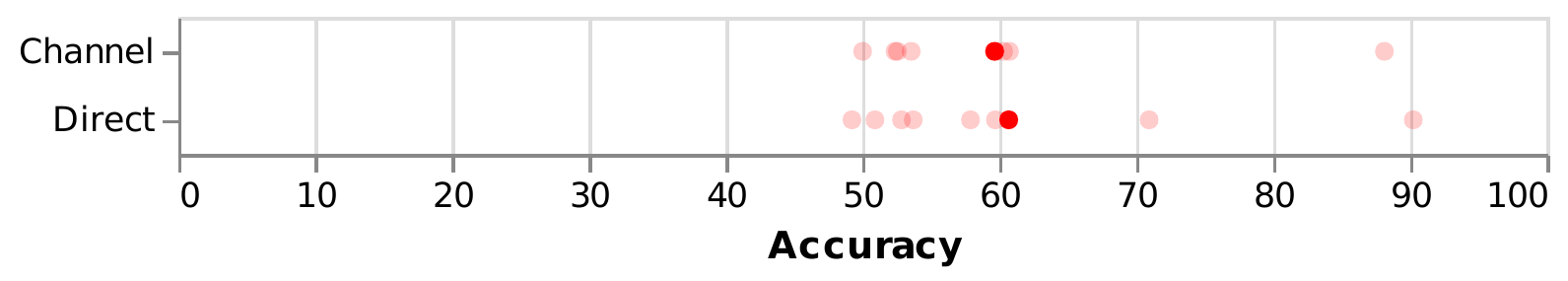}
        \end{subfigure}
        \hfill
        \caption{GPT-J}
    \end{subfigure}
    \caption{The effect of switching the ICL inference method from \textit{Direct} to \textit{Channel}. Employing the Noisy Channel method improves insensitivity while improving the overall ICL performance.}
    \label{fig:when-channel}
\end{figure}

\begin{figure}[t]
    \centering
    \begin{subfigure}{0.45\textwidth}
        \centering
        \begin{subfigure}{\textwidth}
            \centering
            \includegraphics[width=\textwidth]{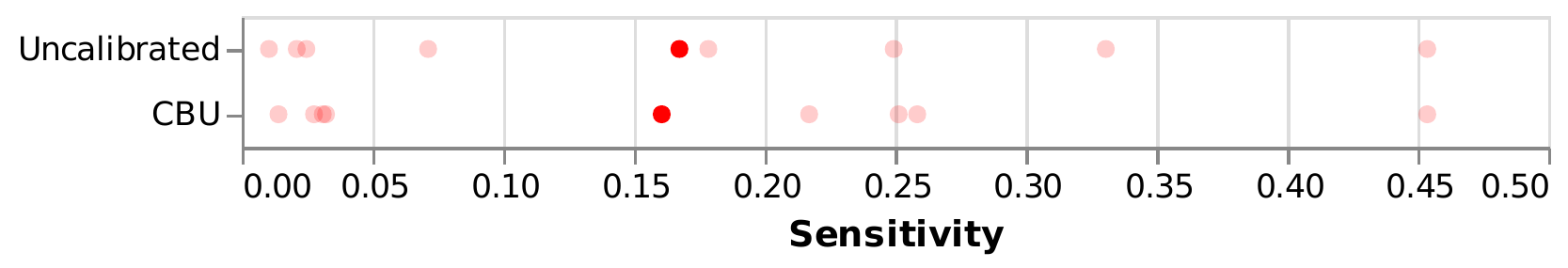}
        \end{subfigure}
        \hfill
        \begin{subfigure}{\textwidth}
            \centering
            \includegraphics[width=\textwidth]{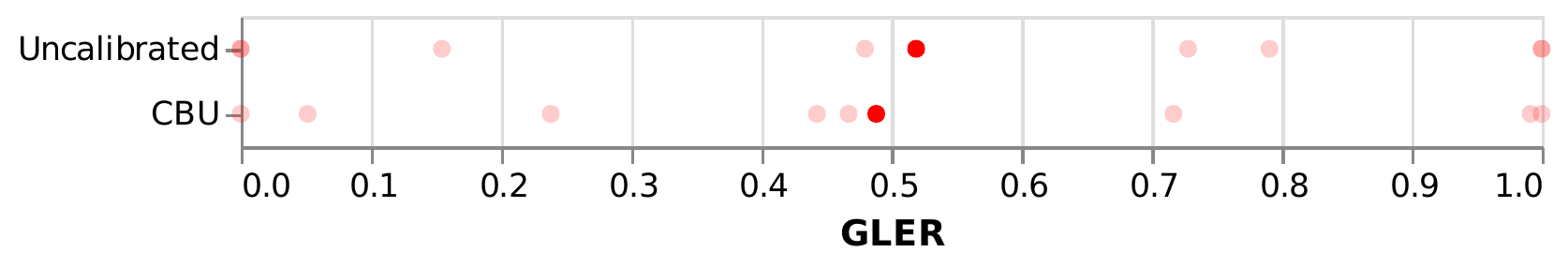}
        \end{subfigure}
        \hfill
        \begin{subfigure}{\textwidth}
            \centering
            \includegraphics[width=\textwidth]{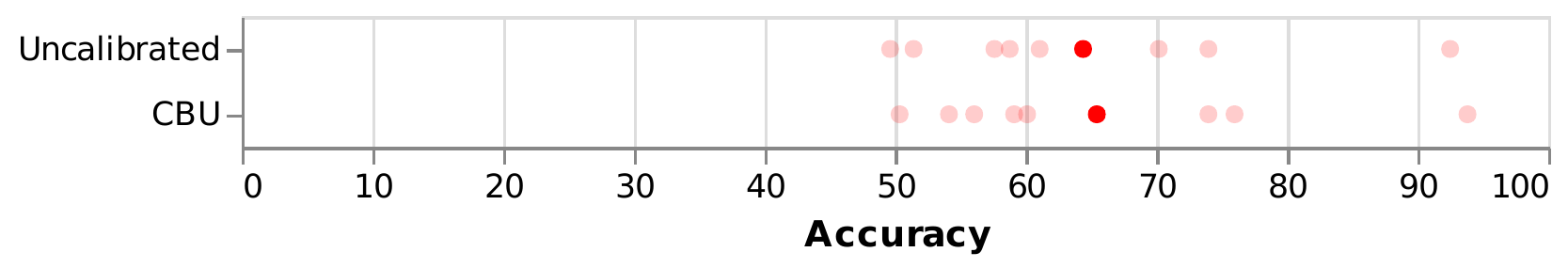}
        \end{subfigure}
        \caption{GPT-NeoX}
    \end{subfigure}
    \centering
    \begin{subfigure}{0.45\textwidth}
        \centering
        \begin{subfigure}{\textwidth}
            \centering
            \includegraphics[width=\textwidth]{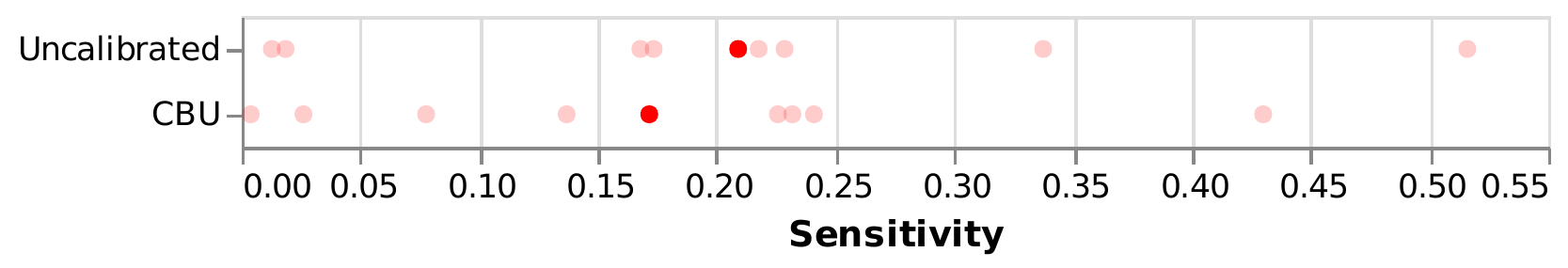}
        \end{subfigure}
        \hfill
        \begin{subfigure}{\textwidth}
            \centering
            \includegraphics[width=\textwidth]{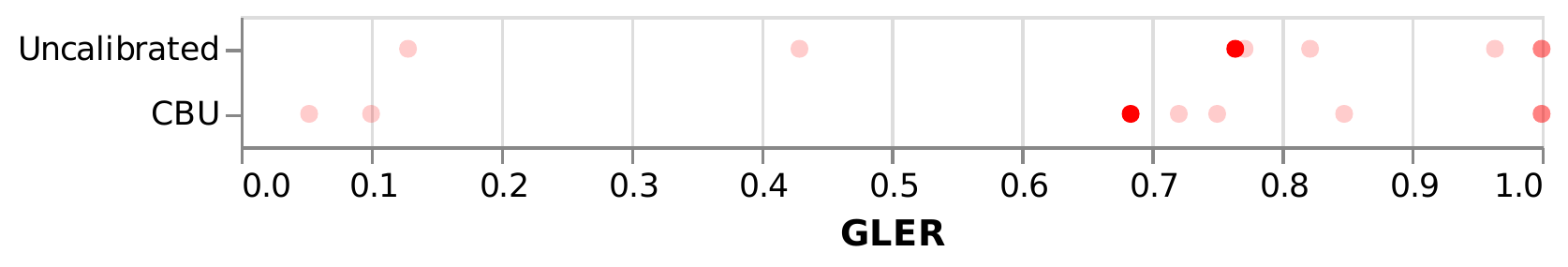}
        \end{subfigure}
        \hfill
        \begin{subfigure}{\textwidth}
            \centering
            \includegraphics[width=\textwidth]{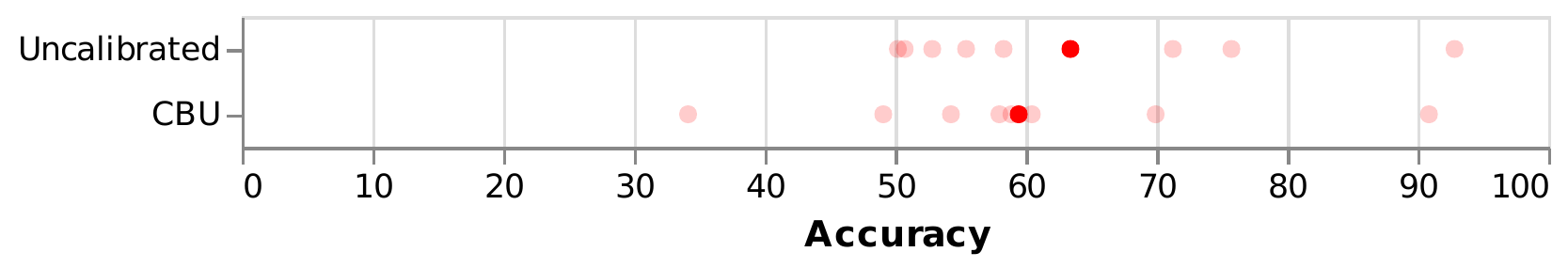}
        \end{subfigure}
        \caption{GPT-J}
    \end{subfigure}
    \caption{The effect of applying Calibrate Before Use (CBU) \cite{zhao2021calibrate}. Label sensitivity decreases but the ground-truth label accuracy improves, making CBU ideal for sensitivity reduction. This trend is more apparent in the larger GPT-variant, GPT-NeoX (20B).}
    \label{fig:when-cbu}
\end{figure}

In-context learning, as first proposed by \citet{brown2020language}, is a straightforward parameter-free approach, where the downstream task of interest is expressed as natural text demonstrations and used to conditionally generate from a language model. Recently, \citet{min2021noisy} proposed Noisy Channel (denoted as \textit{Channel}) that exploits the language generation capability of language models for discriminative tasks using the Bayes' Theorem. We compare the two ICL methods on all three (sensitivity, GLER, and the ground-truth label ICL accuracy) measures.

Results (Figure \ref{fig:when-channel}) show that Channel reduces the label-sensitivity on average compared to the original Direct method while maintaining the Accuracy on similar levels. The label insensitivity effect is observed in both GPT-NeoX and GPT-J.

Another recent advance in ICL, namely Calibrate Before Use (CBU), involves calibrating the output likelihood of the word tokens that correspond to the labels \cite{zhao2021calibrate}.
We conduct the same set of experiments with CBU applied and report all three metrics. As shown in Figure \ref{fig:when-cbu}, the calibration technique reduces the label sensitivity while generally improving the ICL performance on both GPT-J and GPT-NeoX. Applying CBU can be an effective way to reduce label sensitivity while not sacrificing the performance.

\subsection{Prompt Templates}

\label{sec:when-template}

\begin{figure*}[t]
    \centering
    \begin{subfigure}{\textwidth}
        \begin{subfigure}{0.3\textwidth}
        \centering
        \includegraphics[width=\textwidth]{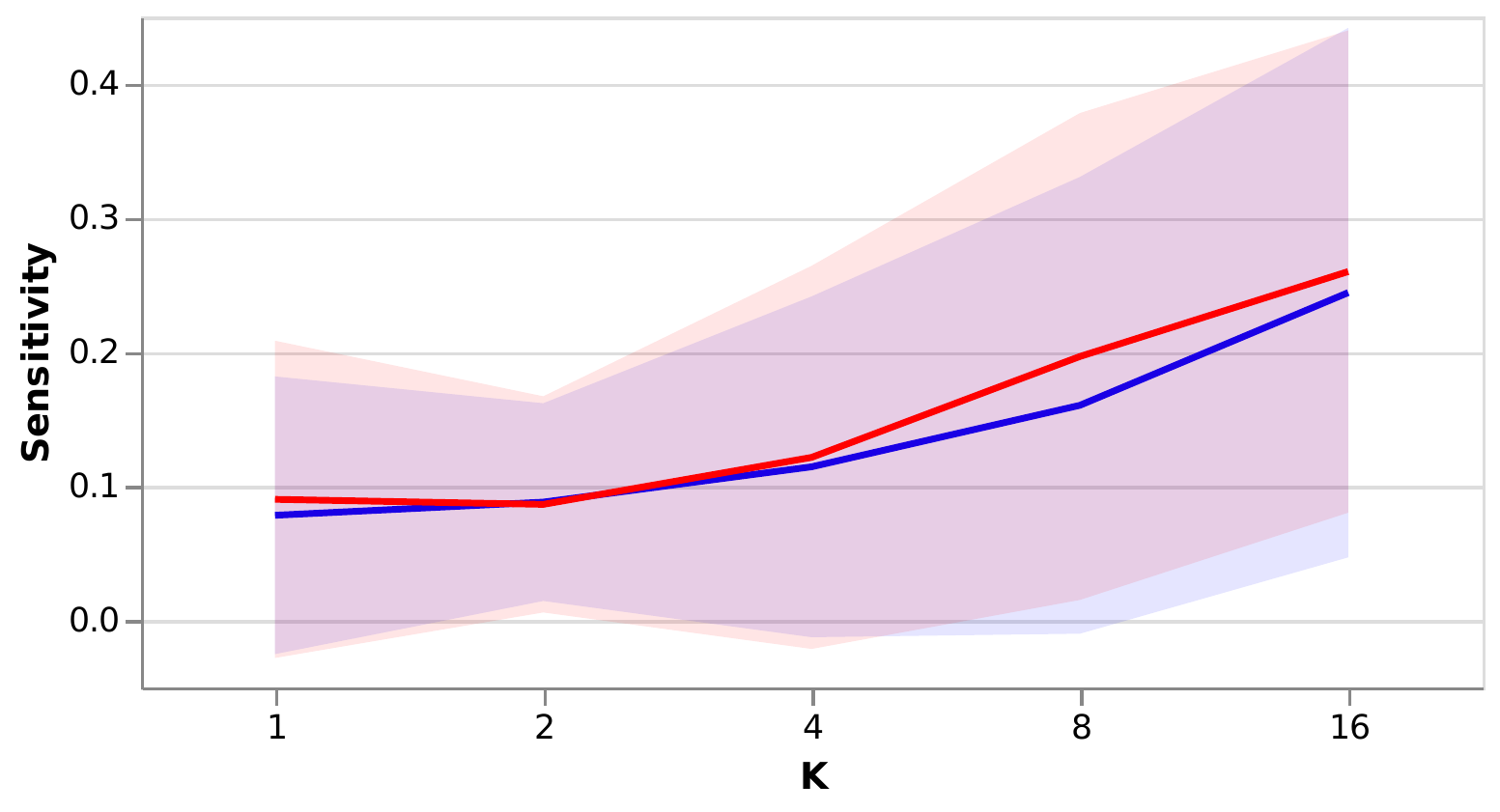}
        \end{subfigure}
        \hfill
        \begin{subfigure}{0.3\textwidth}
            \centering
            \includegraphics[width=\textwidth]{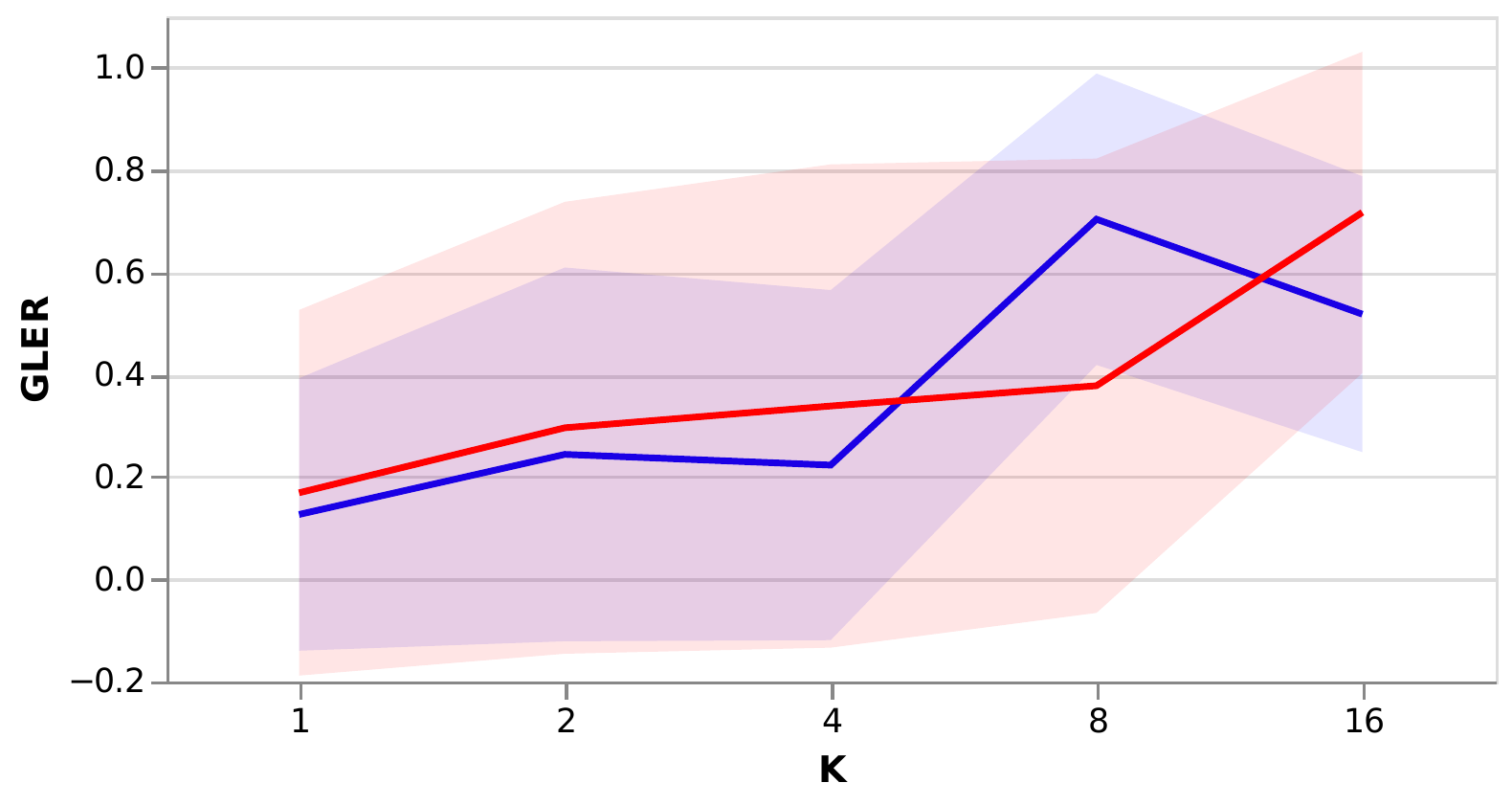}
        \end{subfigure}
        \hfill
        \begin{subfigure}{0.3\textwidth}
            \centering
            \includegraphics[width=\textwidth]{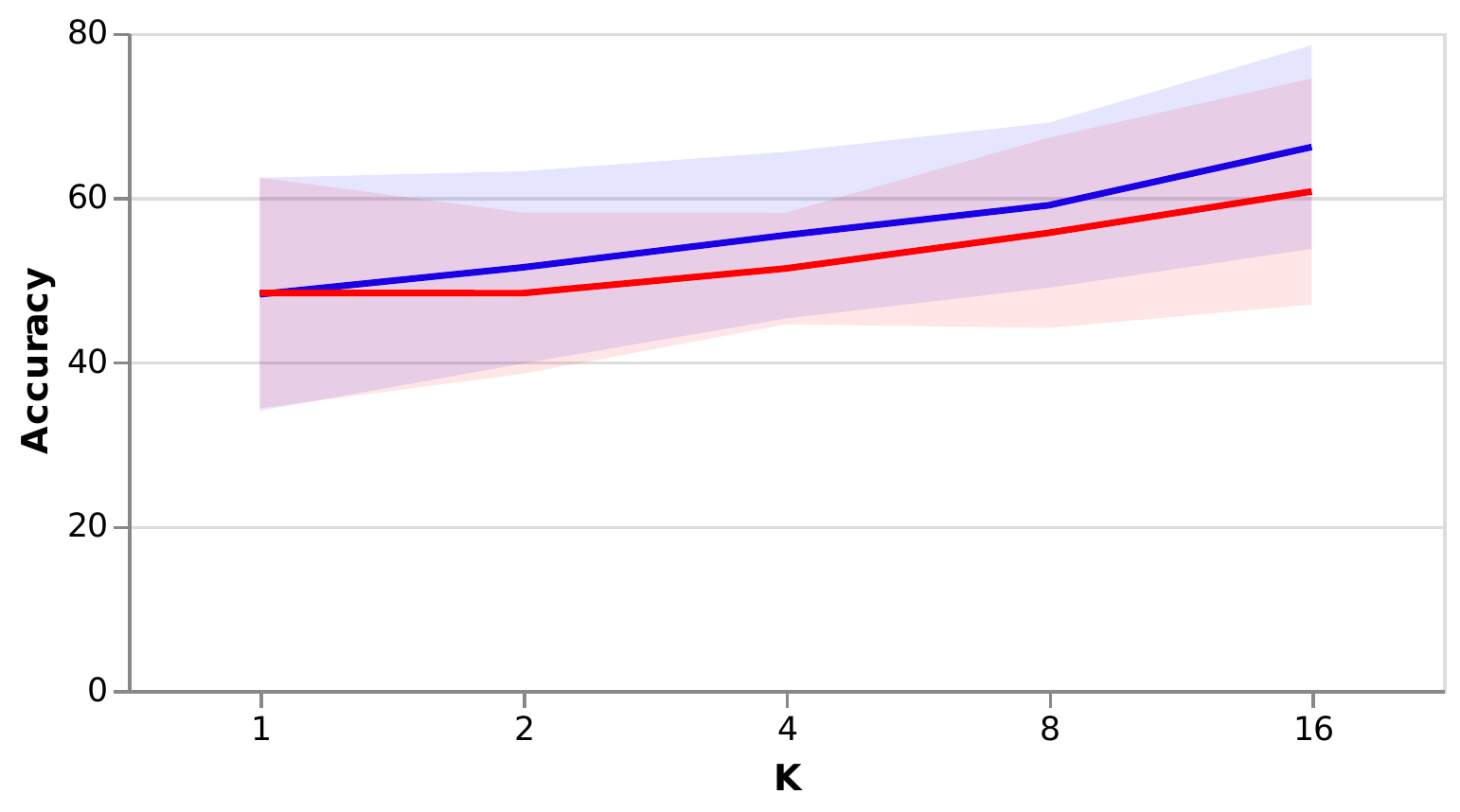}
        \end{subfigure}
        \caption{Number of in-context examples}
        \label{fig:when-examples}
    \end{subfigure}
    \begin{subfigure}{\textwidth}
        \centering
        \begin{subfigure}{0.3\textwidth}
            \centering
            \includegraphics[width=\textwidth]{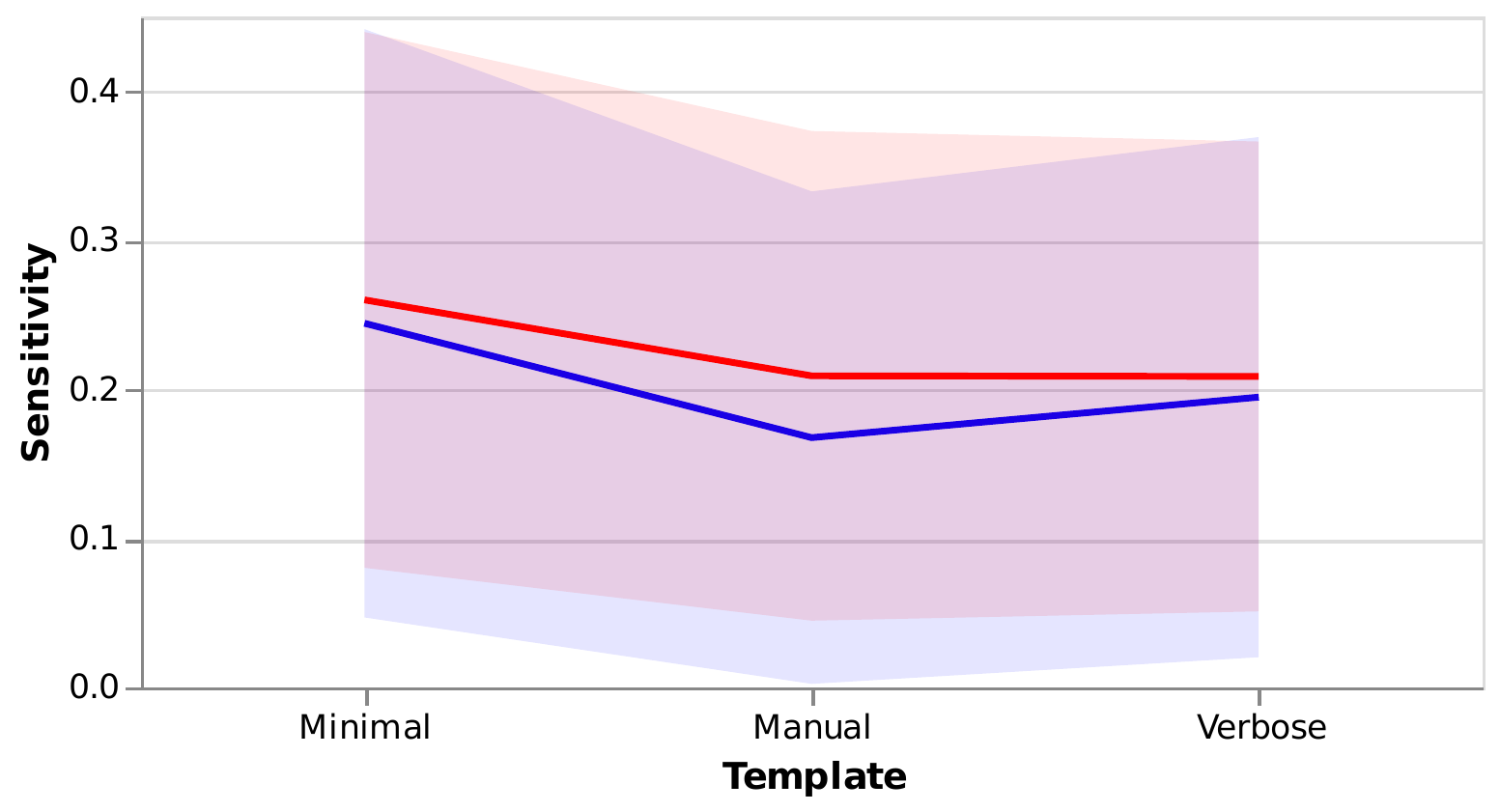}
        \end{subfigure}
        \hfill
        \begin{subfigure}{0.3\textwidth}
            \centering
            \includegraphics[width=\textwidth]{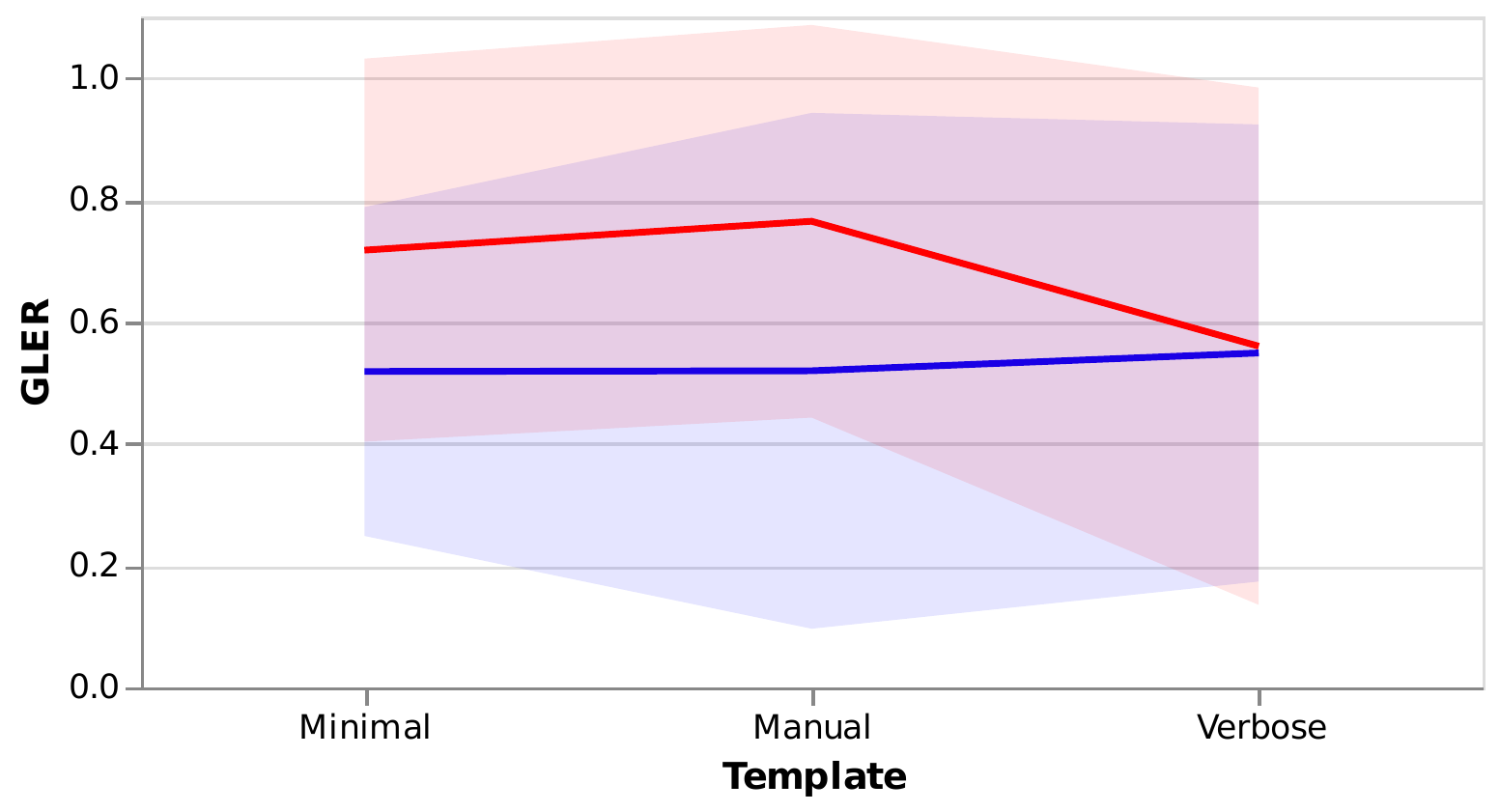}
        \end{subfigure}
        \hfill
        \begin{subfigure}{0.3\textwidth}
            \centering
            \includegraphics[width=\textwidth]{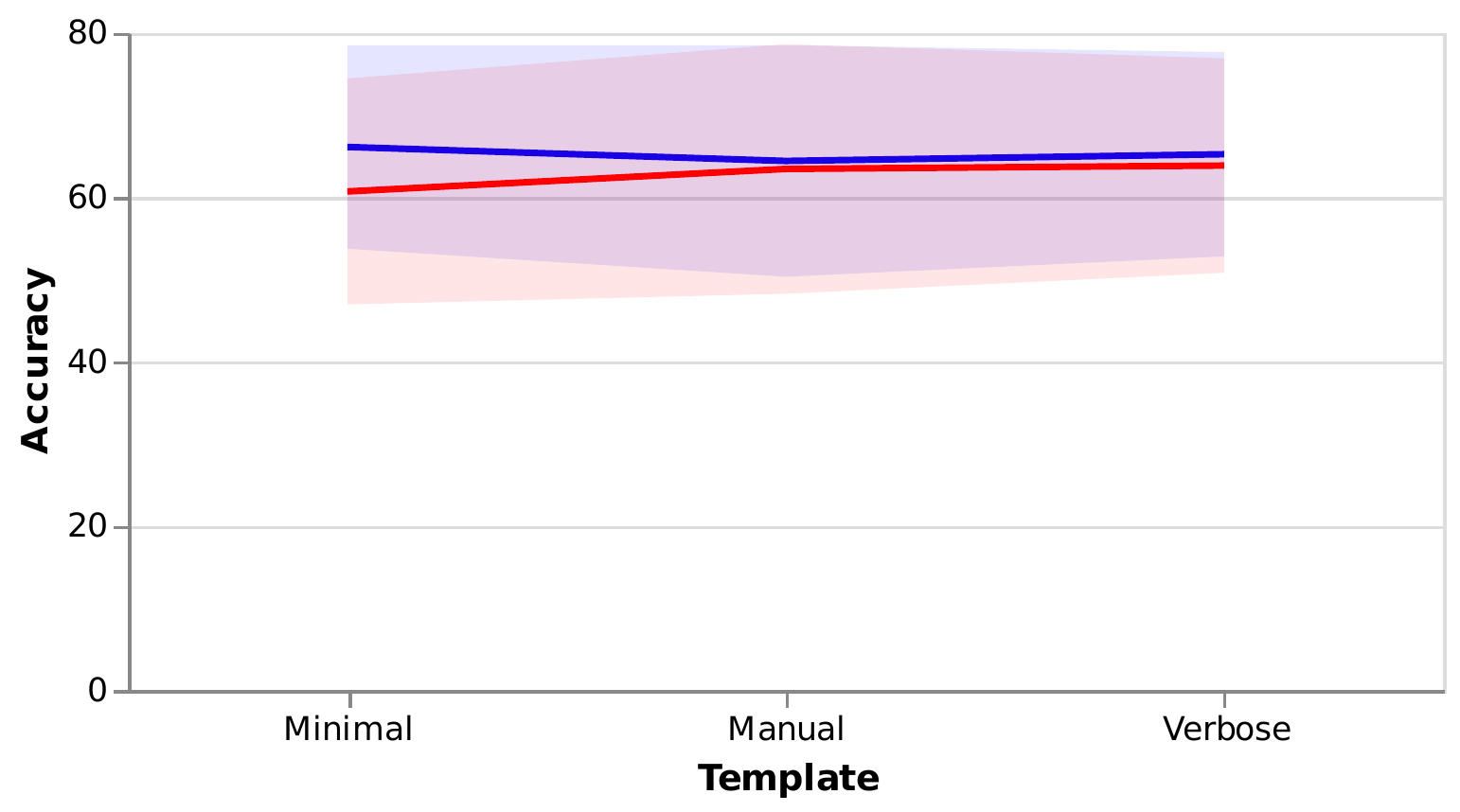}
        \end{subfigure}
        \caption{Prompt template verbosity}
        \label{fig:when-template}
    \end{subfigure}
    \caption{Results for varying prompt sizes and prompt verbosity. The sensitivity, impact ratio, and final ground-truth label performance are all positively correlated with the number of prompt examples. For template verbosity, the sensitivity and the impact ratio decreases with the increase in versbosity, but the performance does not deterioriate. Results for GPT-NeoX (20B) are colored blue, while GPT-J (6B) is colored red.}
\end{figure*}

Various design choices in in-context prompt templates have significant impact on the downstream ICL performance \cite{reynolds2021prompt}. A well-designed and verbose prompt template (e.g., a prompt with detailed description of the task) could allow in-context label demonstrations to have relatively less impact on ICL, thereby reducing the label-demonstration sensitivity. 

This section mainly explores (1) the number of in-context examples and (2) the level of task description details. To quantify the impact of the number of in-context examples, we conduct the same set of experiments with varying number of in-context examples, ranging from 1 to 16. Results (Figure \ref{fig:when-examples}) unsurprisingly show that the number of prompt examples is positively linked to all three metrics. Although sensitivity rises with the number of examples, this is due to the final ICL performance and the impact of ground-truth labels improving with more demonstration examples. 

We also hypothesize that the level of task details contained in the prompt template also serves to relatively weaken the label demonstration impact. Results in Figure \ref{fig:when-template} confirm our hypothesis.

\subsection{Model Sizes}
\label{sec:when-model}

\begin{figure*}[t]
    \centering
    \begin{subfigure}{0.3\textwidth}
        \centering
        \includegraphics[width=\textwidth]{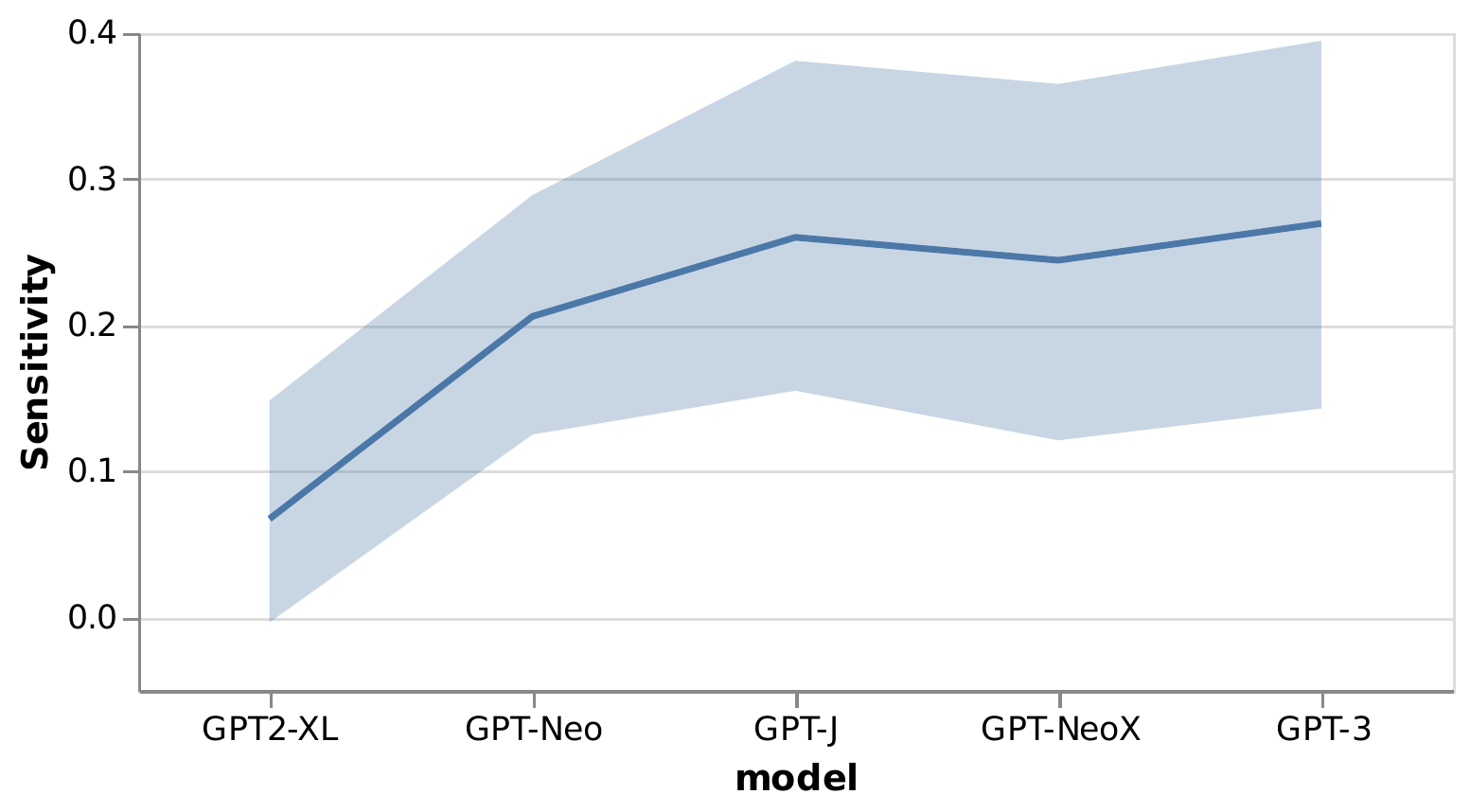}
    \end{subfigure}
    \hfill
    \begin{subfigure}{0.3\textwidth}
        \centering
        \includegraphics[width=\textwidth]{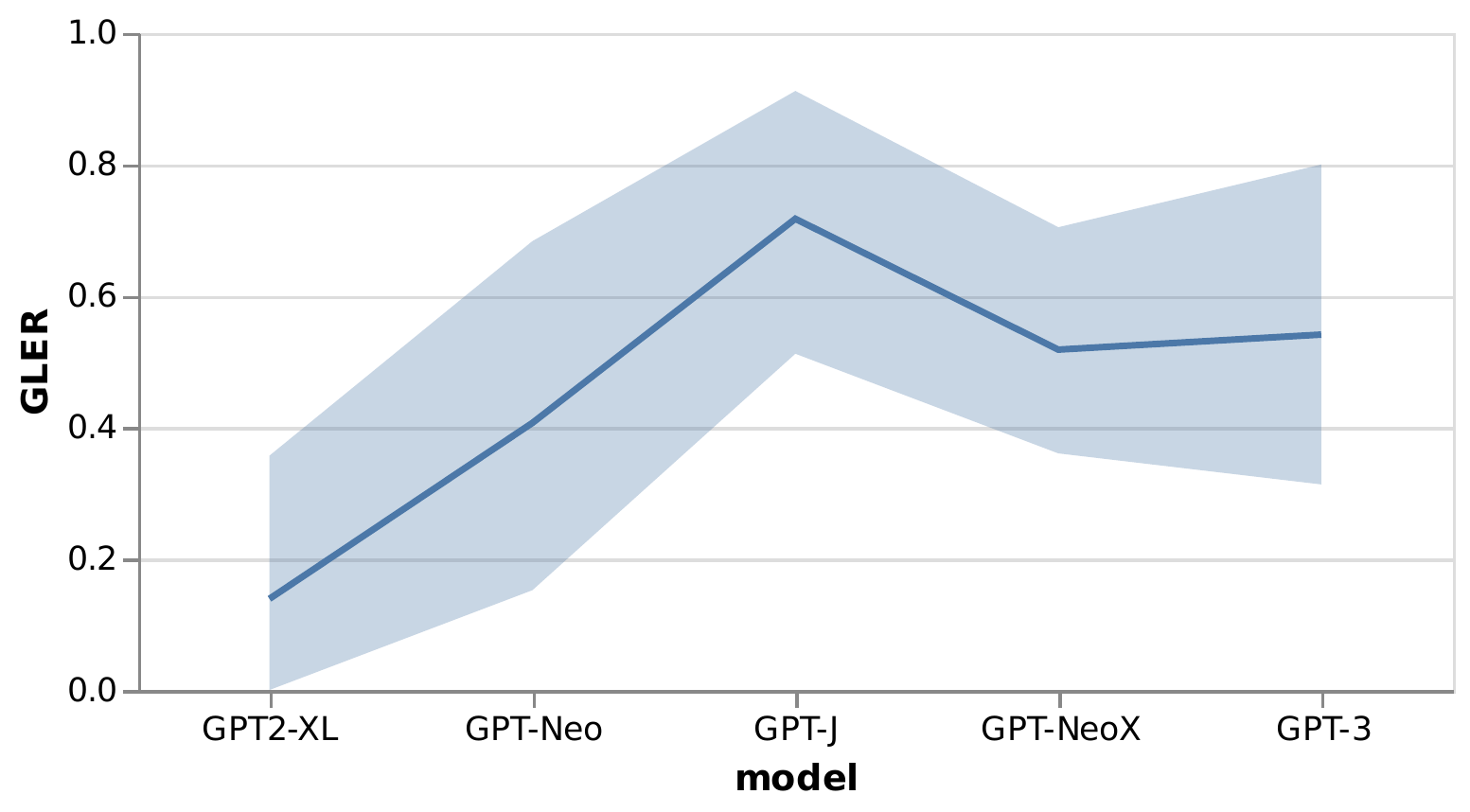}
    \end{subfigure}
    \hfill
    \begin{subfigure}{0.3\textwidth}
        \centering
        \includegraphics[width=\textwidth]{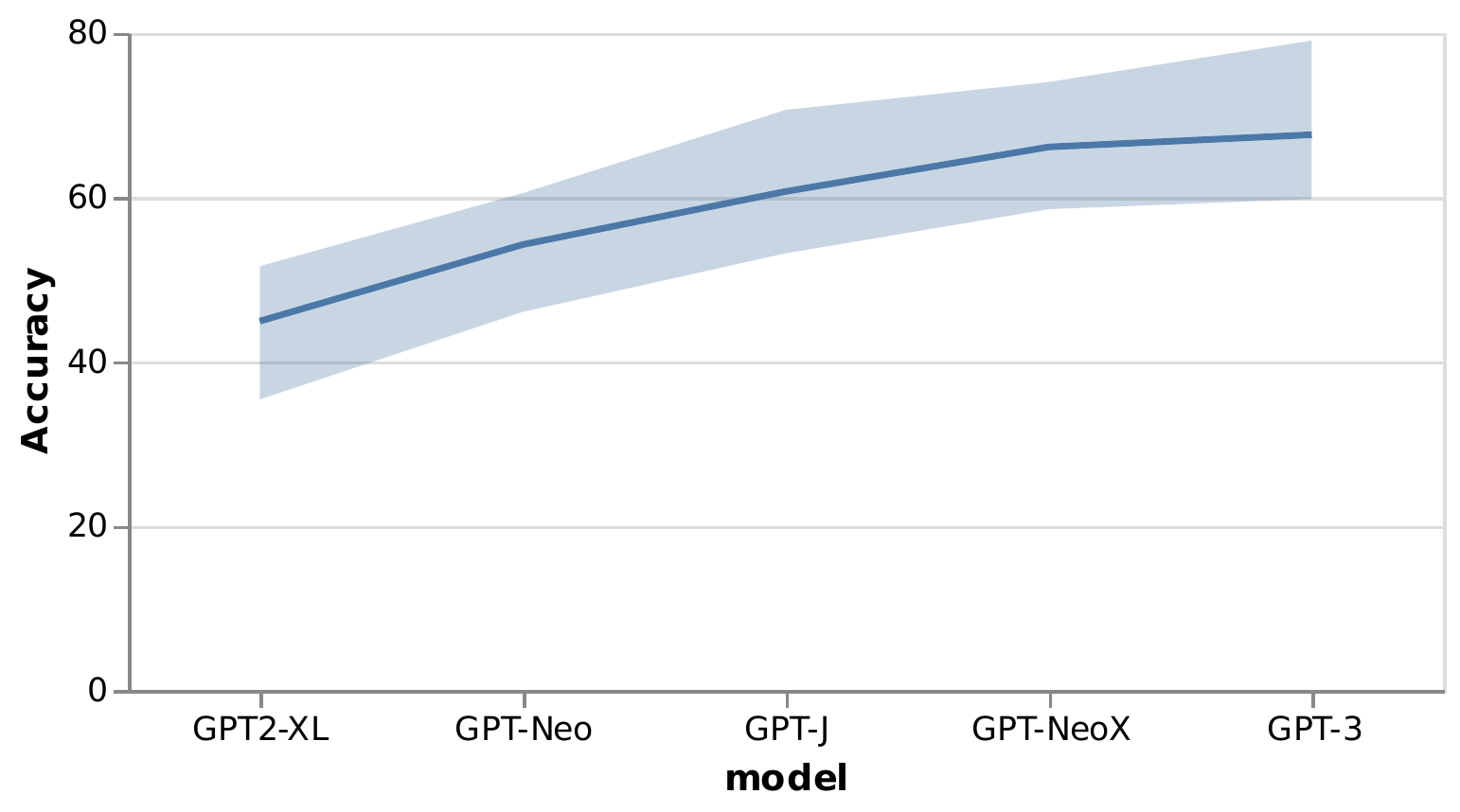}
    \end{subfigure}
    \caption{Comparison of sensitivity, GLER, and the ground-truth label ICL accuracy across different model sizes, ranging from GPT2-XL (1.5B) to GPT-3 (175B). Sensitivity and GLER plateau from the six-billion scale (GPT-J) while the ground-truth label performance continues to improve as the model size scales up.}
    \label{fig:model-size}
\end{figure*}

The scale of the language model could influence how susceptible the model is to label noises within input-label demonstrations. The larger the model is, the more prior knowledge the model could leverage to reduce label sensitivity. To study whether this is the case, we analyze five different sizes of GPT-style language models, ranging from GPT-2 XL to GPT-3\footnote{Note that the general trend along the model scale persists with mixed language model architectures, as reported by \citet{srivastava2022beyond}}. The choice of models and the corresponding number of parameters are listed in Figure \ref{fig:model-size}. Results show that sensitivity is generally correlated with the model size, but we also observe a plateauing phenomenon after the GPT-J 6B scale. However, the results on the ICL performance with ground-truth label demonstrations shows that the performance scales well beyond the 6B mark, 

\section{Discussion}
\label{sec:diss}

\begin{figure}[t]
    \centering
    \includegraphics[width=0.50\textwidth]{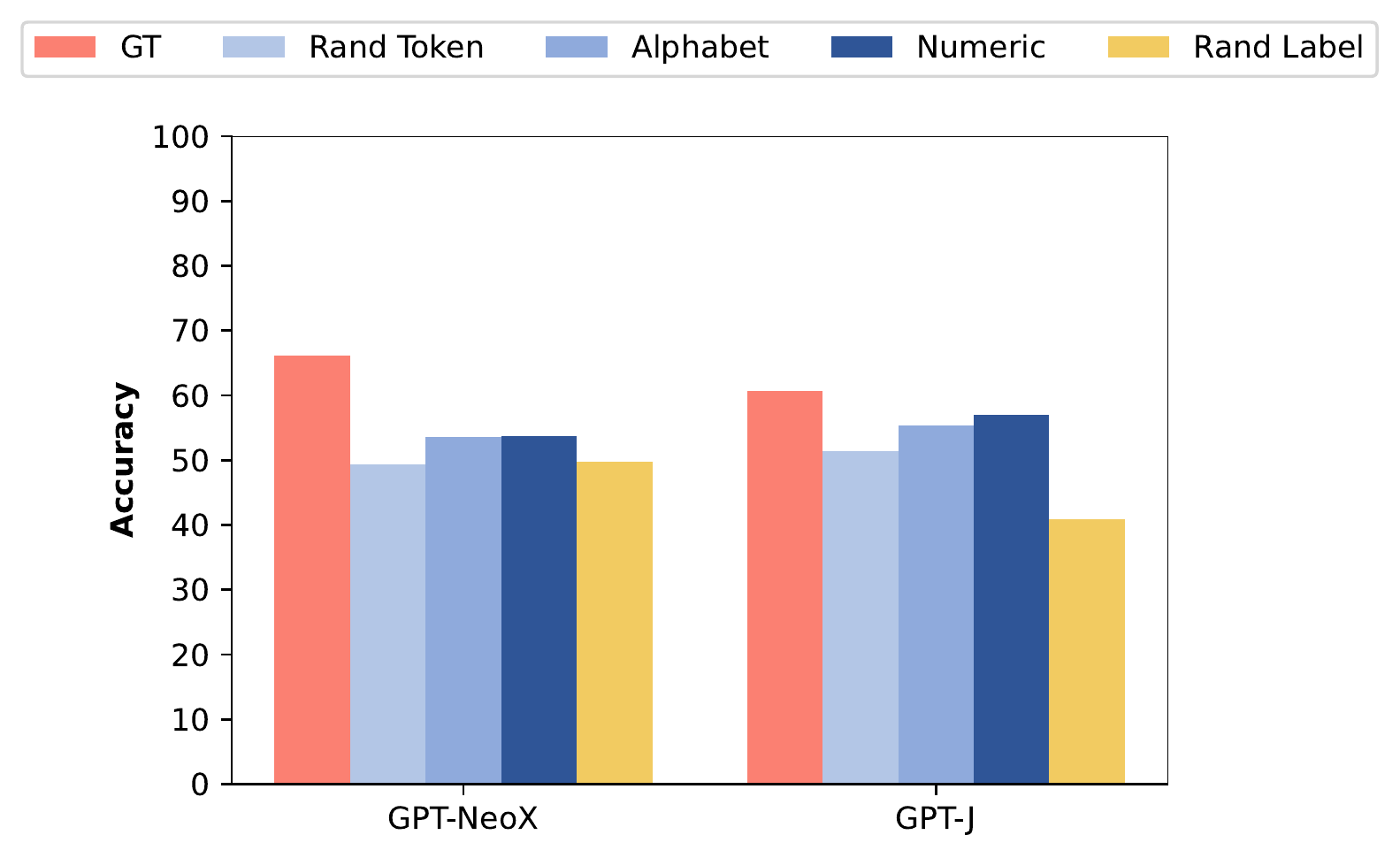}
    \caption{The results of ``label prior-free'' experiments (on 8 text classification datasets), where we control the prior information of the class labels. Here, the labels are replaced with tokens that are unrelated to the label semantics while still maintaining the input-label mappings. The replacement tokens include alphabet tokens, numeric tokens, and random word tokens from the language model's word space (``rand token''). The baselines obtained from the ground-truth labels and random labels are denoted as ``GT'' and ``rand label'' respectively. Results strongly suggest that language models are still able to utilize input-label demonstrations without access to label priors.}
    \label{fig:diss-prior}
\end{figure}

\begin{figure}
    \centering
    \begin{subfigure}{0.45\textwidth}
        \centering
        \begin{subfigure}{\textwidth}
            \centering
            \includegraphics[width=\textwidth]{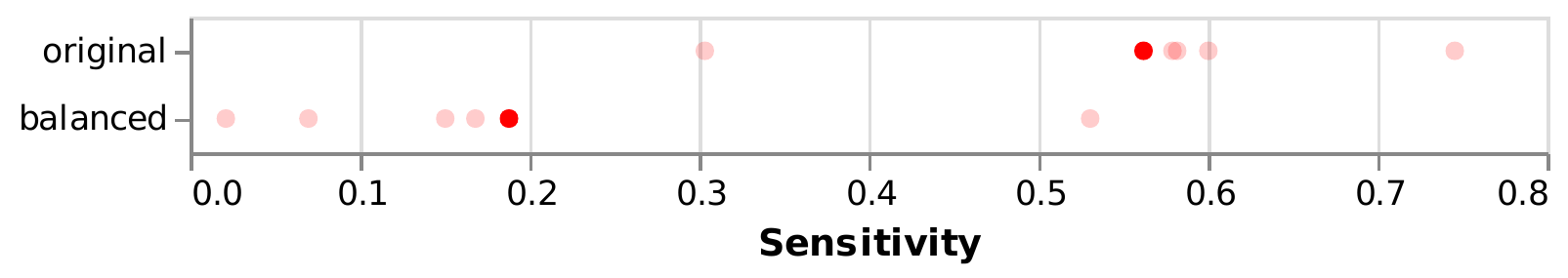}
        \end{subfigure}
        \hfill
        \begin{subfigure}{\textwidth}
            \centering
            \includegraphics[width=\textwidth]{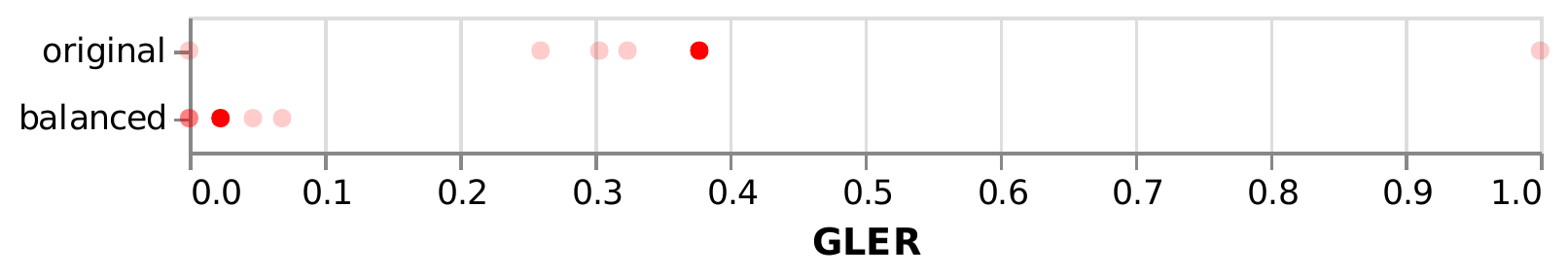}
        \end{subfigure}
        \hfill
        \begin{subfigure}{\textwidth}
            \centering
            \includegraphics[width=\textwidth]{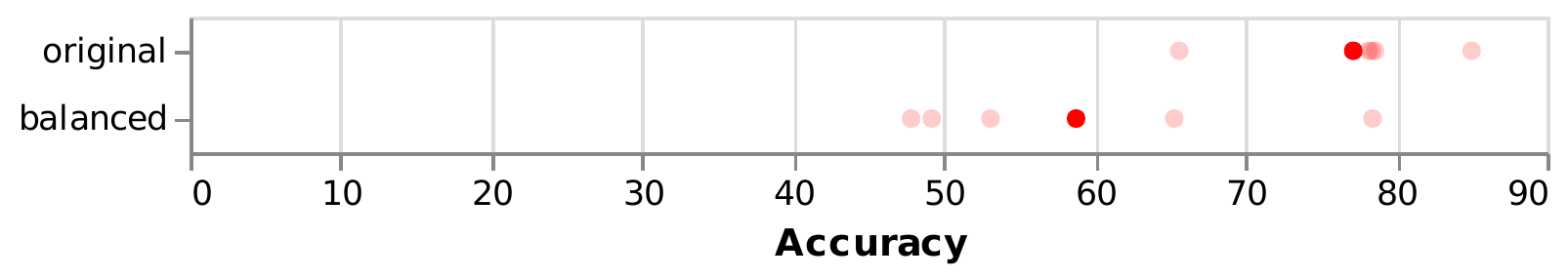}
        \end{subfigure}
        \caption{GPT-NeoX}
    \end{subfigure}
    \centering
    \begin{subfigure}{0.45\textwidth}
        \centering
        \begin{subfigure}{\textwidth}
            \centering
            \includegraphics[width=\textwidth]{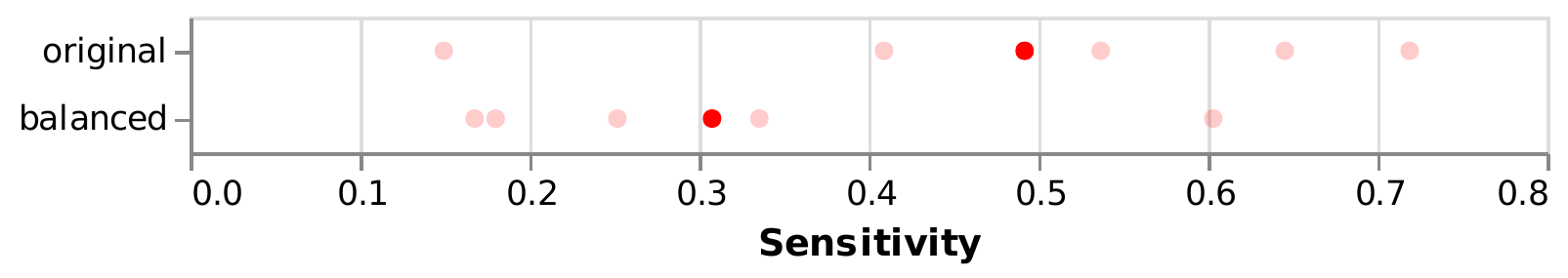}
        \end{subfigure}
        \hfill
        \begin{subfigure}{\textwidth}
            \centering
            \includegraphics[width=\textwidth]{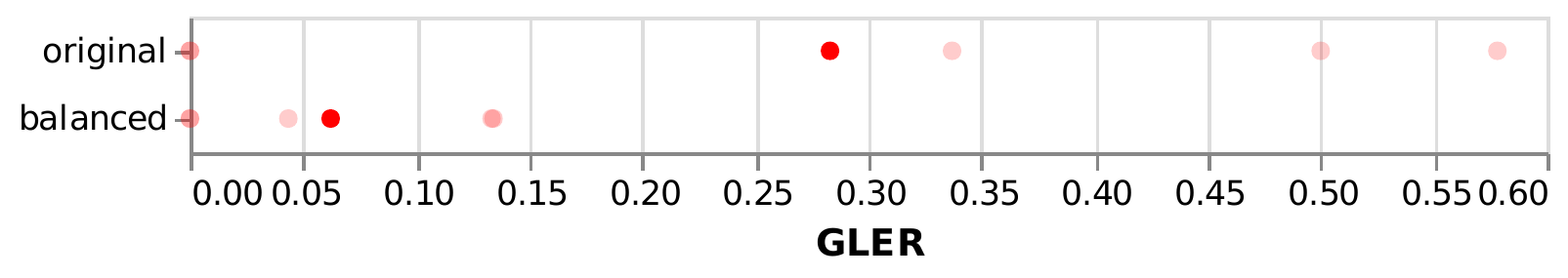}
        \end{subfigure}
        \begin{subfigure}{\textwidth}
            \centering
            \includegraphics[width=\textwidth]{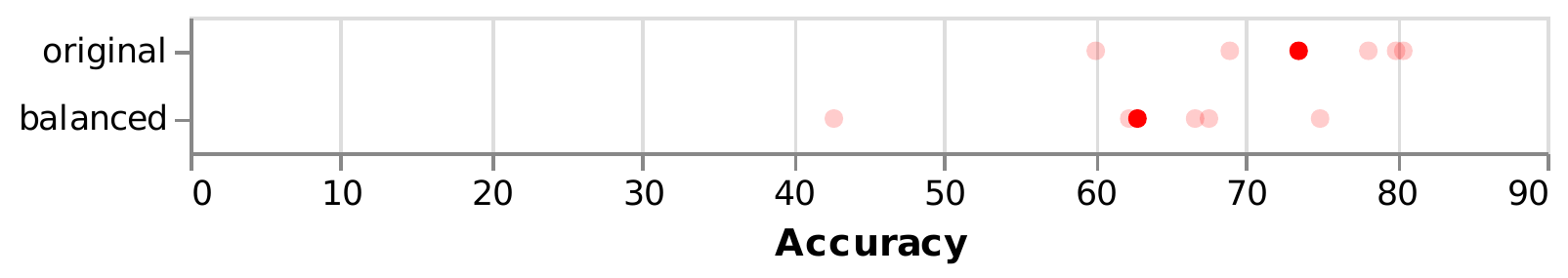}
        \end{subfigure}
        \hfill
        \caption{GPT-J}
    \end{subfigure}
    \caption{The effect of using label balanced demonstrations in 5 imbalanced datasets. Employing the balanced demonstrations degrade all metrics due to the distributional shift in label demonstrations. However, sensitivity is still significant which supports the importance of input-label demonstrations.}
    \label{fig:diss-balance}
\end{figure}

This section provides additional evidence that the demonstration of ground-truth labels can be more important than the previous finding suggests and that existing interpretation of the experimental results may have been obfuscated by the entanglement of various aspects of demonstrations. 

\subsection{The Complementary Relationship between Input-label Correspondence and Label-space Demonstrations}

Input-label correspondence is just one of the aspects of possible in-context label demonstrations, the others including label-space demonstration. However, it is unclear whether label-space and input-label correspondence can complement each other in the absence of explicit demonstration of the other. For example, pretrained language models may be able to deduce sentence-sentiment mappings from the mentions of sentiment labels alone through inductive bias.

Prior work \citep{min2022rethinking} showed significant performance degradation in the absence of both aspects of label demonstration, but the results beg the question: could the significant degradation have been caused by \textit{complete lack of label demonstration}? To find out, we conduct additional ablation studies to study the performance under the demonstration of input-label pairings but not of the explicit label space which we call \textit{prior-free} label experiments. 

Specifically, we study the case where class labels are replaced with \textit{prior-free} labels while maintaining the correspondence between the input and the labels. For example, ``positive'' and ``negative'' labels in sentiment analysis can be replaced with ``0'' and ``1'' labels respectively, which do not reveal the information about the labels themselves. However, language models can still capture mild label-associations in abstract symbols through inductive bias \citep{ouyang2022training}. To diversity ``prior-free'' choices, we consider (1) random tokens from the language model's word space, (2) alphabets, and (3) numerical labels\footnote{We exclude ``0'' since it is often associated with the state of nil}.

As shown in \ref{fig:diss-prior}, results on \textit{prior-free} labels outperform that of the random labels (with random input-label mappings), indicating that language models are capable of capturing the input-label correspondence even in the absence of label-space demonstrations. Among the prior-free results, we note that the alphabetical and numerical labels outperform random-token labels. This could be explained by the fact that, since random word tokens may introduce unintended biases through misleading association with unrelated word semantics, abstract labels provide better prior-free environment.

\subsection{Change in label distribution may result the higher sensitivity.} 
The distribution of labels in demonstration is one of the critical factor for the prediction \cite{zhao2021calibrate}. When data imbalance exists, corrupting the labels cause distributional shift which may lead performance change regardless of the input-label mappings. High sensitivity in imbalanced dataset may be due to this unintentional distributional shift. To analyze the impact of distributional shift, we conducted additional experiments using label balanced demonstrations for imbalanced dataset (hate\_speech18, ethos-race, ethos-national\_origin, ethos-religion). 

As shown in \ref{fig:diss-balance}, using balanced demonstrations degrade the performance and sensitivity when compared to demonstrations sampled from data distributions which supports our suspicion. On the other hand, average sensitivity are 0.189 and 0.308 (for GPT-NeoX and GPT-J respectively) even in balanced demonstrations setting which supports the importance of input-label demonstrations.

\section{Related Work}
\label{sec:related}

As the scale of language models becomes larger \cite{rae2021scaling,chowdhery2022palm,smith2022using,thoppilan2022lamda}, fine-tuning becomes prohibitively expensive due to the space and time complexities. As an alternative, in-context learning (ICL) \cite{brown2020language} has shown to be an effective parameter-free learning strategy by prompting language models with task-specific prompt templates. Since then, a plethora of works has investigated both the properties of the learning mechanism \cite{schick2021s,reynolds2021prompt,kim2021changes,zhao2021calibrate,lu2021fantastically,min2022rethinking}. Although numerous efficient fine-tuning strategies have been proposed in the past \cite{prefixtuning,LoRA,lester2021power}, the absence of an explicit training step in ICL has enabled it to retain its own class of adapting large-scale language models.

\section{Conclusion and Future Work}

In this work, we took a closer look at how input-label relationships affect the in-context learning performance.
To quantitatively analyze the impact of input-label mappings in in-context learning, we proposed novel metrics, GLER and input-label sensitivity.
Through extensive experiments, we found that the integrity of the input-label mapping is a crucial factor in performing ICL.
We also conducted ablation studies to reveal various conditions that allow ICL to improve insensitivity towards label corruptions (while still maintaining a healthy performance).
For future work, based on the current findings, we will investigate whether we could exploit data augmentation for extremely low-resource situations for ICL.

\section*{Limitations}

\paragraph{PLMs are over sensitive to the choice of prompts.} 
As it is widely known that performance of the PLMs is highly sensitive to the choice of the prompts \cite{brown2020language,lu2021fantastically,zhao2021calibrate}. Prompt engineering to find the optimal prompt was not feasible considering the amount of datasets and settings that we experimented. The findings from this work may differ depending on the choice of prompts. However, to minimize this limitations the templates and prompts are adopted from well studied previous works as much as possible. 

\paragraph{Ground-truth label demonstrations are just one piece of the puzzle.} According the full analysis from \citet{min2022rethinking}, other components of demonstrations not covered in this paper (e.g., input-space demonstrations) exhibit even stronger impacts on ICL. Although our experiments were designed to analyze solely the impact of input-label correspondence, disentangling diverse aspects of demonstrations is highly difficult as mentioned in section \ref{sec:diss}. Other factors such as label distribution may have unexpectedly influenced the results.

\paragraph{Huggingface Implementation.} We use Huggingface implementation of GPT-NeoX. To our knowledge, current version of GPT-NeoX in Huggingface under performs when compared to the original implementations from \citet{gpt-neox-20b}.

\section*{Acknowledgement}

This work was mainly supported by SNU-NAVER Hyperscale AI Center and partly supported by Institute of Information \& communications Technology Planning \& Evaluation (IITP) grant funded by the Korea government(MSIT) [No.2020-0-01373, Artificial Intelligence Graduate School Program (Hanyang University), No.2021-0-01343,
Artificial Intelligence Graduate School Program (Seoul National
University)].
Last but not least, we would like to express gratitude to Yejin Choi for the insightful discussions and feedback.

\bibliographystyle{acl_natbib}
\bibliography{emnlp2022-gtlm}
\newpage

\clearpage
\appendix

\section{Details on Our Experimental Settings}
\label{app:counter-example}
\label{app:experiment_detail}

\subsection{Model}
We mainly experiment with GPT-Neox 20B \cite{gpt-neox-20b} and GPT-J 6B \cite{gpt-j} which are publicly released, decoder-only, dense LMs. However, in Section \ref{sec:when-model} we also include GPT2-XL 1.5B \cite{radford2019language}, GPT-Neo 2.7B \cite{gpt-neo}, GPT-3 175B \cite{brown2020language}. 

\subsection{Full Dataset}
We evaluate on 17 text classification datasets covering diverse tasks including sentiment analysis, paraphrase detection, natural language inference, hate speech detection and diverse domains including science, social media, finance, and more. All datasets are from Huggigface datasets \cite{lhoest2021datasets}. Full list and details about the datasets are provided in Table \ref{tab:data_detail}.

As mentioned in Section \ref{sec:looking-analyses-sensitivity}, sensitivity highly depends on relative performance. In order to effectively capture correlation between sensitivity and diverse factors in Section \ref{sec:when}, we evaluate on subset of 8 datasets, datasets with high relative performance, in Section \ref{sec:when}. 8 datasets include glue-sst2, glue-rte, super\_glue-cb, trec, financial\_phrasebank, medical\_questions\_pairs, sick, and tweet\_eval-hate. Due to limited resources, we only run experiments on 6 datasets in Section \ref{sec:when-model}.

\subsection{Metric}
We use accuracy as our primary metric. Accuracy is commonly used metric in multi-class classification which intuitively show how well the model performs. F1 score takes into account how the data is distributed thus it is useful when you have data with imbalance classes. However, F1 is less intuitive since it measures the \textit{trade-off} between precision and recall. Moreover, F1 score can vary regarding the averaging method in multi-class classification. 

\subsection{Template}
We use 3 types of templates regarding engineering cost and verbosity of templates. First, as a baseline template we used minimal template following \cite{ye2021crossfit, min2022rethinking}. We use minimal template throughout the paper. For ablation \ref{sec:when-template}, we also evaluate manual templates and Verbose template. Templates are adopted from prior works  \cite{brown2020language, zhao2021calibrate, min2022rethinking, bach2022promptsource} if possible. Details and examples regarding the templates are in Table \ref{tab:template-detail}.
Additionally, for Section \ref{sec:when-icl} CBU experiment we use Manual template as the baseline since in our preliminary experiments, applying CBU in Minimal template degrade the performance in some cases. 

Even though we use the same minimal template as \citet{min2022rethinking}, there are minor difference in dataset-specific implementation of data preprocessor. (e.g., input sentences of glue\_mrpc dataset used in \citet{min2022rethinking} have prefix "sentence1: ") Therefore, LMs may have slightly different behavior with same the dataset.

\subsection{Other details}
\label{app:other-details}
Unless otherwise specified, we use k = 16 examples as demonstrations which are sampled at uniform from the training data. We run all experiments 5 times using different seeds. Due to limited resources, we only run experiments once for GPT-3. For all models expect for GPT-3, we used implementation and models from Huggingface transformers library \cite{wolf-etal-2020-transformers}. For GPT-3 we used OpenAI API, assuming that model "davinci" is GPT-3 175B. When calculating the probability of label tokens, we do not normalize the score by the length of the tokens unlike in \citet{min2022rethinking}. Our implementation is available at https://github.com/juny116/ICL-DeepSpeed.

\subsection{Corrupting input-label mapping}
To see the detail impact of the ground truth input-label mapping, we revisit the experiments from \citet{min2022rethinking}
Specifically, we replace fix amount of correct labels to incorrect labels in demonstrations and compare the end task performance.

\begin{itemize}
    \item \textbf{No demonstrations} is a zero-shot prediction made via $argmax_{y\in C} P(y|x)$, where $x$ is the test input and $C$ is a small discrete set of possible labels. Verbalizers are used for mapping tokens to class.
    \item \textbf{Demonstrations w/ \boldsymbol{$a\%$} correct labels} consist $k \times a / 100$ correct pairs and $k \times (1 - a/100)$ incorrect pairs where $(0 \leq a \leq 100)$. A concatenation of k input-label pairs where $a\%$ labels are correct is used to make a prediction via $argmax_{y\in C} P(y|x_1,y_1,...,x_k,y_k,x)$. 
    \item \textbf{Demonstrations w/ random label} is formed with replacing correct labels to random labels that are randomly sampled at uniform from $C$. Since the labels are sampled at uniform from $C$, the distribution of labels in demonstration may change from sampled inputs.
    \item \textbf{Demonstrations w/ shuffled label} is formed with randomly shuffling correct labels to other labels within the sampled k inputs. The distribution of labels in demonstration does not change from sampled inputs. 
    \item \textbf{Majority class baseline} is a ratio of majority class within the test data. Since there are some datasets that have distributional imbalance, this can be a good indicator of how well the in-context learning is working.
\end{itemize}

\begin{table}[t]
    \centering
    \resizebox{0.99\columnwidth}{!}{%
        \begin{tabular}{lrrr}
        \toprule
        Dataset & Train & Eval & Class \\ \midrule
        glue-sst2 \cite{socher2013recursive} & 67,349 & 872 & 2 \\
        glue-rte \cite{dagan2005pascal} & 2,490 & 277 & 2 \\
        glue-mrpc \cite{dolan2005automatically} & 3,668 & 408 & 2 \\
        glue-wnli \cite{levesque2012winograd} & 635 & 71 & 2 \\
        super\_glue-cb \cite{de2019commitmentbank} & 250 & 56 & 3 \\
        trec \cite{voorhees2000building} & 5,452 & 500 & 5 \\
        financial\_phrasebank \cite{malo2014good} & 1,181 & 453 & 3 \\
        poem\_sentiment \cite{sheng2020investigating} & 843 & 105 & 3 \\
        medical\_questions\_pairs \cite{mccreery2020effective} & 2,438 & 610 & 2 \\
        sick \cite{marelli2014sick} & 4,439 & 495 & 3 \\
        hate\_speech18 \cite{de2018hate} & 8,562 & 2,141 & 4 \\
        ethos-national\_origin \cite{mollas2020ethos} & 346 & 87 & 2 \\
        ethos-race \cite{mollas2020ethos} & 346 & 87 & 2 \\
        ethos-religion \cite{mollas2020ethos} & 346 & 87 & 2 \\
        tweet\_eval-hate \cite{barbieri2020tweeteval} & 9,000 & 1,000 & 2 \\
        tweet\_eval-stance\_atheism \cite{barbieri2020tweeteval} & 461 & 52 & 3 \\
        tweet\_eval-stance\_feminist \cite{barbieri2020tweeteval} & 597 & 67 & 3 \\ \bottomrule
        \end{tabular}
    }
    \caption{Datasets used for the experiment.}
    \label{tab:data_detail}
\end{table}

\begin{table*}[t]
    \centering
    \resizebox{0.99\textwidth}{!}{%
        \begin{tabular}{lll}
        \toprule
        Dataset & Manual Template & Verbalizer \\ \midrule
        glue-sst2 & \begin{tabular}[c]{@{}l@{}}\textcolor{blue}{Review:} a smile on your face\\ \textcolor{blue}{Sentiment:}\end{tabular} & negative, positive \\
        meddical\_questions\_pairs & \begin{tabular}[c]{@{}l@{}}The DVD-CCA then appealed to the state Supreme Court .\\ \textcolor{blue}{The question is:} The DVD CCA appealed that decision to the U.S. Supreme Court . \textcolor{blue}{True or False?}\\ \textcolor{blue}{answer:}\end{tabular} & False, True \\
        glue-rte & \begin{tabular}[c]{@{}l@{}}Oil prices fall back as Yukos oil threat lifted\\ \textcolor{blue}{The question is:} Oil prices rise. \textcolor{blue}{True or False?}\\ \textcolor{blue}{answer:}\end{tabular} & True, False \\
        super\_glue-cb & \begin{tabular}[c]{@{}l@{}}That was then, and then's gone. It's now now. I don't mean I 've done a sudden transformation.\\ \textcolor{blue}{The question is:} she has done a sudden transformation \textcolor{blue}{True or False?}\\ \textcolor{blue}{answer:}\end{tabular} & True, False, Not sure \\
        trec & \begin{tabular}[c]{@{}l@{}}\textcolor{blue}{Question:} How can I find a list of celebrities ' real names ?\\ \textcolor{blue}{Type:}\end{tabular} & \begin{tabular}[c]{@{}l@{}}description, entity,\\ expression, human,\\ number, location\end{tabular} \\
        sick & \begin{tabular}[c]{@{}l@{}}The young boys are playing outdoors and the man is smiling nearby\\ \textcolor{blue}{The question is:} The kids are playing outdoors near a man with a smile \textcolor{blue}{True or False?}\\ \textcolor{blue}{answer:}\end{tabular} & True, Not sure, False \\
        tweet\_eval-hate & \begin{tabular}[c]{@{}l@{}}\textcolor{blue}{Tweet:} Hundreds of Syrian refugees return home from Lebanon - ABC News\\ \textcolor{blue}{Sentiment:}\end{tabular} & favor, against \\ \bottomrule
        \end{tabular}
    }
    \resizebox{0.99\textwidth}{!}{%
        \begin{tabular}{lll}
        \toprule
        Dataset & Verbose Template & Verbalizer \\ \midrule
        glue-sst2 & \begin{tabular}[c]{@{}l@{}}\textcolor{blue}{Question : Is the following review positive or negative?} a smile on your face\\ \textcolor{blue}{Answer:}\end{tabular} & negative, positive \\
        meddical\_questions\_pairs & \begin{tabular}[c]{@{}l@{}}\textcolor{blue}{Question: Does the following two sentences mean the similar thing? True or False?} \\ The DVD-CCA then appealed to the state Supreme Court .\\ The DVD CCA appealed that decision to the U.S. Supreme Court .\\ \textcolor{blue}{Answer:}\end{tabular} & False, True \\
        glue-rte & \begin{tabular}[c]{@{}l@{}}\textcolor{blue}{Question: Does the first sentence entails the second sentence? True or False? } \\ Oil prices fall back as Yukos oil threat lifted\\ Oil prices rise. \\ \textcolor{blue}{Answer:}\end{tabular} & True, False \\
        super\_glue-cb & \begin{tabular}[c]{@{}l@{}}\textcolor{blue}{Question: Does the first sentence entails the second sentence? True, False, or Neither?} \\ That was then, and then's gone. It's now now. I don't mean I 've done a sudden transformation.\\  she has done a sudden transformation \\ \textcolor{blue}{Answer:}\end{tabular} & True, False, Neither \\
        trec & \begin{tabular}[c]{@{}l@{}}\textcolor{blue}{Question: Which category best describes the following sentence?} \\ How can I find a list of celebrities ' real names ?\\ \textcolor{blue}{Answer:}\end{tabular} & \begin{tabular}[c]{@{}l@{}}description, entity,\\ expression, human,\\ number, location\end{tabular} \\
        sick & \begin{tabular}[c]{@{}l@{}}\textcolor{blue}{Question: Does the first sentence entails the second sentence? True, False, or Not sure?} \\ The young boys are playing outdoors and the man is smiling nearby\\ The kids are playing outdoors near a man with a smile \\ \textcolor{blue}{Answer:}\end{tabular} & True, Not sure, False \\
        tweet\_eval-hate & \begin{tabular}[c]{@{}l@{}}\textcolor{blue}{Question: Does the tweet convey the author’s hatred towards something or someone? True or False?} \\ Hundreds of Syrian refugees return home from Lebanon - ABC News\\ \textcolor{blue}{Answer:}\end{tabular} & True, False \\ \bottomrule
        \end{tabular}
    }
    \caption{Examples of Manual and Verbose templates. Texts in blue are manual templates.}
    \label{tab:template-detail}
\end{table*}

\section{Full Results}
\label{app:full-results}
    Full experiment results on 17 datasets with GPT-NeoX are in Table \ref{tab:full-neox} and results on 17 datasets with GPT-NeoX are in Table \ref{tab:full-j}.

\section{More Results on the Sensitivity vs Task Difficulty Plot}

Figure \ref{fig:sensitivity-difficulty} shows scatter plots of sensitivities of 17 datasets against the corresponding task difficulties measured using the relative performance with respect to accuracy and F-1 scores. The Direct approach is colored in orange and the Channel approach is colored in blue. The dashed vertical line indicates a neutral performance level where there is no difference with the random baselines. The best-fit linear lines show a general trend of increasing sensitivity with less task difficulty. Low sensitivity is strongly related to high task difficulty. Also, the Channel approach helps in alleviating hyper-sensitivity towards task difficulty.
\label{app:task-difficulty}

\begin{figure}[t]
    \centering
    \begin{subfigure}{0.4\textwidth}
        \centering
        \includegraphics[width=\textwidth]{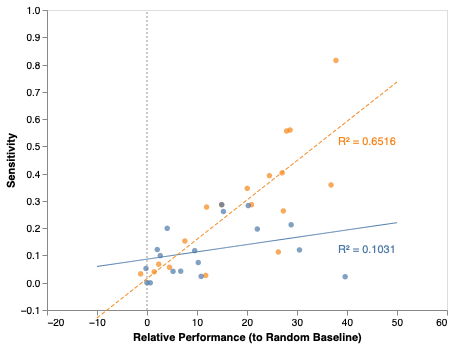}
        \caption{Accuracy}
    \end{subfigure}

    \begin{subfigure}{0.4\textwidth}
        \centering
        \includegraphics[width=\textwidth]{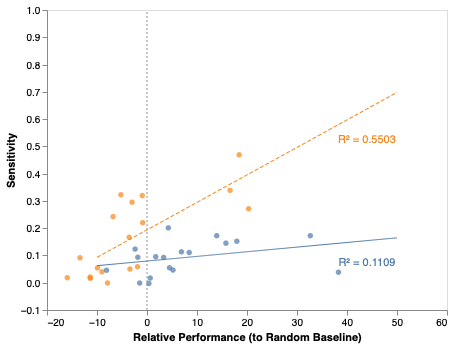}
        \caption{F1-score}
    \end{subfigure}

    \caption{Scatter plots of sensitivities of 17 datasets against the corresponding task difficulties measured using the relative performance with respect to each metrics.}
    \label{fig:sensitivity-difficulty}
\end{figure}

\section{Label-Correctness Correlation} 

\begin{figure}[t]
    \centering
    \includegraphics[width=0.45\textwidth]{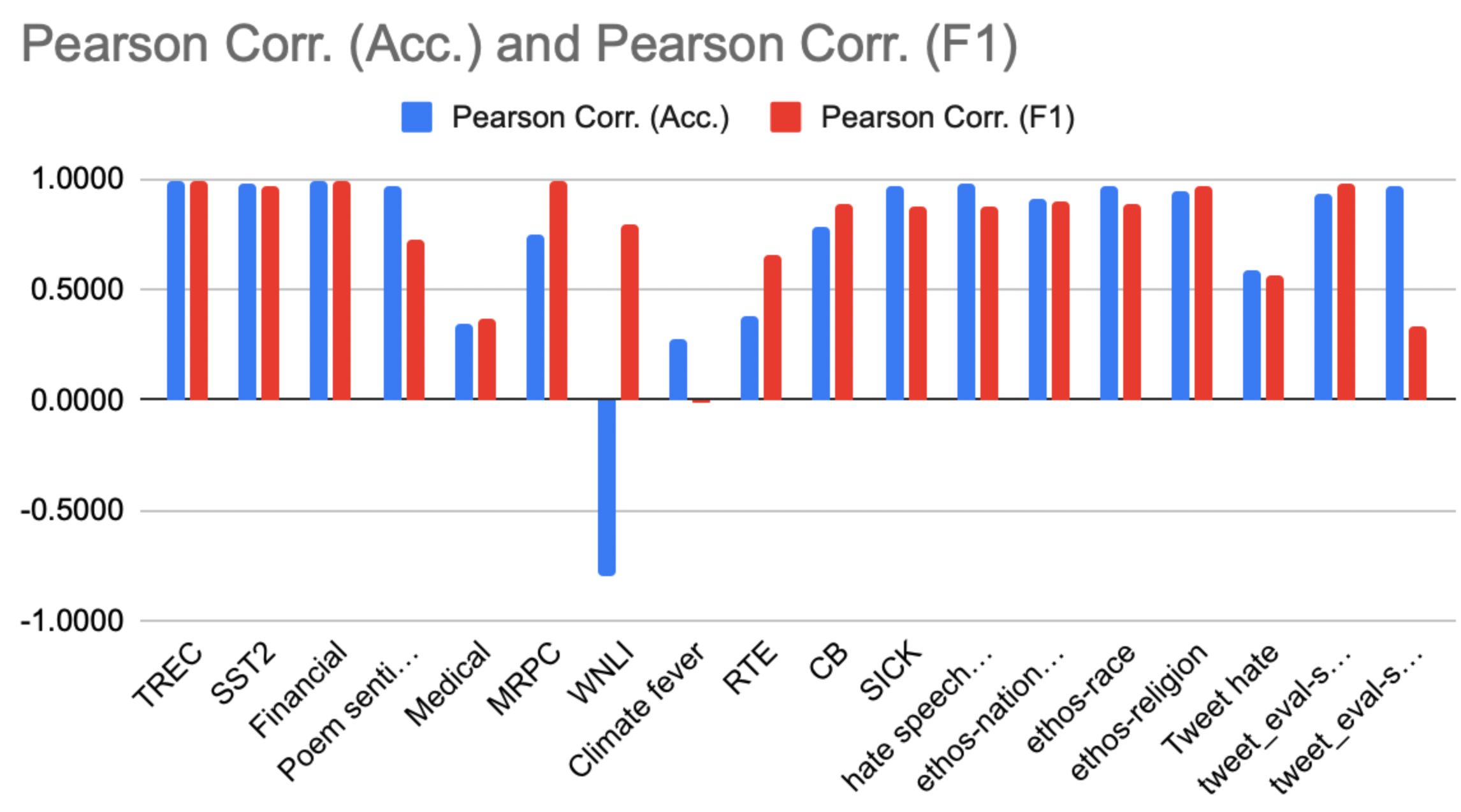}
    \caption{Pearson correlation analysis on all 18 tasks. A strong positive correlation is observed for all tasks and metrics, except for outliers.}
    \label{fig:correlation}
\end{figure}

The first step of understanding the interaction between performance and input-label demonstration is quantifying the correlation between the two variables. Although we considered this metric as one of the foundation quantifying measures, we omit the analyses results due to space constraints. The Pearson correlation analysis on GPT-J and the Direct approach (Figure \ref{fig:correlation}) shows that the label-correctness correlation is strong (i.e. larger than 0.9) for most tasks on all performance measures. The macro-average correlation across 18 tasks is 0.895 with a p-value of 0.057, strongly supporting the linkage.

\begin{table*}[ht]
    {\tiny
    \begin{tabular}{lccccccccc}
    \toprule
    \multicolumn{1}{c}{Dataset} & Metric & no demo & 0\% & 25\% & 50\% & 75\% & 100\% & random label & shuffled label \\ \midrule
    \multirow{2}{*}{\begin{tabular}[c]{@{}l@{}}glue\\ sst2\end{tabular}} & Accuracy & 74.54 & 70.67$\pm7.40$ & 70.880$\pm7.40$ & 70.80$\pm10.61$ & 78.44$\pm18.24$ & 88.26$\pm5.07$ & 84.93$\pm11.29$ & 84.31$\pm9.43$ \\
     & F1 & 73.94 & 67.34$\pm10.13$ & 66.96$\pm14.15$ & 74.50$\pm13.19$ & 74.78$\pm23.46$ & 88.06$\pm5.33$ & 84.09$\pm13.03$ & 83.59$\pm10.59$ \\ \hline
    \multirow{2}{*}{\begin{tabular}[c]{@{}l@{}}glue\\ rte\end{tabular}} & Accuracy & 52.71 & 54.80$\pm3.49$ & 55.38$\pm5.19$ & 52.42$\pm4.36$ & 55.52$\pm3.78$ & 57.04$\pm7.17$ & 55.88$\pm5.11$ & 56.68$\pm3.56$ \\
     & F1 & 34.52 & 47.52$\pm7.80$ & 48.50$\pm9.19$ & 45.33$\pm5.71$ & 51.80$\pm5.51$ & 48.81$\pm14.17$ & 47.97$\pm10.65$ & 51.43$\pm6.01$ \\ \hline
    \multirow{2}{*}{\begin{tabular}[c]{@{}l@{}}glue\\ mrpc\end{tabular}} & Accuracy & 68.38 & 31.86$\pm1.54$ & 30.54$\pm0.81$ & 32.60$\pm4.55$ & 44.36$\pm16.72$ & 53.58$\pm20.06$ & 35.64$\pm8.54$ & 45.20$\pm18.49$ \\
     & F1 & 40.61 & 26.79$\pm4.62$ & 23.86$\pm0.13$ & 28.24$\pm6.82$ & 34.75$\pm11.75$ & 35.89$\pm10.53$ & 29.79$\pm9.66$ & 35.15$\pm11.24$ \\ \hline
    \multirow{2}{*}{\begin{tabular}[c]{@{}l@{}}glue\\ wnli\end{tabular}} & Accuracy & 56.34 & 50.99$\pm6.86$ & 55.77$\pm6.55$ & 48.17$\pm5.58$ & 45.35$\pm7.41$ & 44.79$\pm5.93$ & 51.83$\pm6.09$ & 48.17$\pm6.33$ \\
     & F1 & 36.04 & 45.10$\pm8.76$ & 50.02$\pm10.19$ & 40.65$\pm9.21$ & 35.99$\pm8.15$ & 32.67$\pm5.76$ & 42.60$\pm8.98$ & 38.89$\pm7.16$ \\ \hline
    \multirow{2}{*}{\begin{tabular}[c]{@{}l@{}}super\_glue\\ cb\end{tabular}} & Accuracy & 8.93 & 21.07$\pm14.58$ & 41.43$\pm6.36$ & 46.79$\pm7.72$ & 54.29$\pm7.32$ & 60.36$\pm11.39$ & 30.00$\pm9.48$ & 49.29$\pm6.51$ \\
     & F1 & 5.56 & 16.38$\pm8.88$ & 31.03$\pm7.53$ & 38.23$\pm4.68$ & 41.73$\pm6.07$ & 49.02$\pm9.35$ & 24.48$\pm5.55$ & 31.34$\pm4.18$ \\ \hline
    \multirow{2}{*}{trec} & Accuracy & 21.20 & 34.84$\pm8.34$ & 41.20$\pm12.44$ & 43.92$\pm8.89$ & 56.84$\pm6.21$ & 67.44$\pm6.04$ & 42.60$\pm10.79$ & 36.16$\pm5.59$ \\
     & F1 & 11.85 & 21.26$\pm6.16$ & 24.85$\pm11.79$ & 31.45$\pm8.37$ & 41.72$\pm6.86$ & 52.98$\pm5.50$ & 25.64$\pm11.94$ & 24.45$\pm3.59$ \\ \hline
    \multirow{2}{*}{\begin{tabular}[c]{@{}l@{}}financial\\ phrasebank\end{tabular}} & Accuracy & 21.85 & 25.03$\pm2.31$ & 37.62$\pm8.53$ & 42.38$\pm14.95$ & 79.51$\pm5.70$ & 80.22$\pm8.58$ & 33.60$\pm8.01$ & 57.57$\pm9.43$ \\
     & F1 & 17.50 & 26.78$\pm8.21$ & 42.88$\pm8.40$ & 47.72$\pm11.33$ & 78.31$\pm4.48$ & 75.72$\pm6.01$ & 41.36$\pm8.81$ & 43.56$\pm10.26$ \\ \hline
    \multirow{2}{*}{\begin{tabular}[c]{@{}l@{}}poem\\ sentiment\end{tabular}} & Accuracy & 21.90 & 35.81$\pm24.35$ & 20.95$\pm2.61$ & 40.95$\pm40.94$ & 59.05$\pm5.75$ & 61.52$\pm8.45$ & 44.19$\pm7.87$ & 50.67$\pm14.37$ \\
     & F1 & 22.62 & 19.25$\pm6.99$ & 18.11$\pm5.99$ & 27.05$\pm14.71$ & 35.84$\pm6.20$ & 35.62$\pm9.43$ & 31.689$\pm8.10$ & 30.87$\pm2.74$ \\ \hline
    \multirow{2}{*}{\begin{tabular}[c]{@{}l@{}}medical\\ questions\_pairs\end{tabular}} & Accuracy & 49.51 & 49.34$\pm1.98$ & 49.34$\pm1.80$ & 48.92$\pm1.05$ & 50.98$\pm1.15$ & 51.93$\pm2.76$ & 51.11$\pm1.62$ & 49.87$\pm0.86$ \\
     & F1 & 33.11 & 38.60$\pm5.63$ & 38.64$\pm7.90$ & 42.25$\pm8.02$ & 43.69$\pm6.19$ & 49.06$\pm3.13$ & 41.49$\pm9.38$ & 40.34$\pm7.87$ \\ \hline
    \multirow{2}{*}{sick} & Accuracy & 56.57 & 32.97$\pm5.58$ & 45.29$\pm7.67$ & 54.06$\pm2.58$ & 55.88$\pm10.26$ & 65.62$\pm4.18$ & 47.31$\pm14.14$ & 50.34$\pm12.12$ \\
     & F1 & 24.96 & 26.14$\pm6.16$ & 33.10$\pm3.44$ & 38.63$\pm6.95$ & 42.19$\pm13.93$ & 49.80$\pm11.15$ & 33.53$\pm13.01$ & 33.95$\pm12.21$ \\ \hline
    \multirow{2}{*}{hate\_speech18} & Accuracy & 89.49 & 13.20$\pm3.70$ & 35.13$\pm18.19$ & 38.30$\pm32.00$ & 77.94$\pm23.46$ & 85.01 $\pm9.90$ & 71.28$\pm21.44$ & 89.49$\pm0.02$ \\
     & F1 & 47.23 & 12.59$\pm4.19$ & 27.56$\pm11.32$ & 39.02$\pm15.88$ & 44.49$\pm6.34$ & 47.26$\pm0.21$ & 42.69$\pm6.81$ & 47.22$\pm0.01$ \\ \hline
    \multirow{2}{*}{\begin{tabular}[c]{@{}l@{}}ethos\\ national\_origin\end{tabular}} & Accuracy & 21.84 & 24.37$\pm9.03$ & 29.20$\pm11.57$ & 47.13$\pm23.03$ & 65.97$\pm21.45$ & 78.39$\pm4.70$ & 63.68$\pm16.12$ & 75.17$\pm11.82$ \\
     & F1 & 22.99 & 22.43$\pm8.12$ & 27.04$\pm9.59$ & 36.04$\pm13.34$ & 46.07$\pm7.46$ & 49.28$\pm5.04$ & 45.55$\pm3.77$ & 52.17$\pm11.04$ \\ \hline
    \multirow{2}{*}{\begin{tabular}[c]{@{}l@{}}ethos\\ race\end{tabular}} & Accuracy & 26.44 & 23.68$\pm3.51$ & 27.59$\pm7.13$ & 48.28$\pm17.66$ & 68.74$\pm7.77$ & 78.16$\pm0.00$ & 61.38$\pm14.35$ & 78.39$\pm0.51$ \\
     & F1 & 242.76 & 20.61$\pm5.15$ & 25.33$\pm8.50$ & 42.13$\pm15.77$ & 45.29$\pm4.75$ & 43.87$\pm0.00$ & 45.90$\pm1.85$ & 44.93$\pm2.36$ \\ \hline
    \multirow{2}{*}{\begin{tabular}[c]{@{}l@{}}ethos\\ religion\end{tabular}} & Accuracy & 21.84 & 22.76$\pm3.19$ & 24.60$\pm5.30$ & 37.24$\pm14.93$ & 58.39$\pm22.49$ & 78.62$\pm2.89$ & 31.38$\pm15.41$ & 69.43$\pm18.44$ \\
     & F1 & 20.57 & 20.67$\pm4.30$ & 22.89$\pm6.45$ & 35.16$\pm14.76$ & 41.11$\pm7.43$ & 44.00$\pm0.92$ & 44.15$\pm1.38$ & 42.34$\pm3.24$ \\ \hline
    \multirow{2}{*}{\begin{tabular}[c]{@{}l@{}}tweet\_eval\\ hate\end{tabular}} & Accuracy & 42.70 & 43.08$\pm2.59$ & 45.48$\pm4.78$ & 47.40$\pm5.81$ & 49.52$\pm4.42$ & 58.00$\pm3.86$ & 52.36$\pm5.41$ & 52.46$\pm7.38$ \\
     & F1 & 29.92 & 35.38$\pm5.73$ & 40.90$\pm7.22$ & 43.30$\pm8.89$ & 43.68$\pm8.22$ & 56.38$\pm5.45$ & 44.70$\pm8.61$ & 50.45$\pm9.29$ \\ \hline
    \multirow{2}{*}{\begin{tabular}[c]{@{}l@{}}tweet\_eval\\ stance\_atheism\end{tabular}} & Accuracy & 53.85 & 18.46$\pm1.72$ & 20.38$\pm3.49$ & 22.31$\pm3.99$ & 21.54$\pm4.59$ & 26.15$\pm10.84$ & 18.85$\pm2.51$ & 22.31$\pm4.63$ \\
     & F1 & 41.50 & 14.51$\pm2.93$ & 18.27$\pm4.14$ & 20.24$\pm4.52$ & 17.77$\pm3.76$ & 22.40$\pm12.80$ & 16.31$\pm4.66$ & 17.52$\pm6.31$ \\ \hline
    \multirow{2}{*}{\begin{tabular}[c]{@{}l@{}}tweet\_eval\\ feminist\end{tabular}} & Accuracy & 49.25 & 28.06$\pm4.99$ & 31.64$\pm5.11$ & 29.25$\pm6.38$ & 30.78$\pm9.12$ & 38.51$\pm5.42$ & 29.96$\pm4.30$ & 35.22$\pm4.79$ \\
     & F1 & 34.97 & 20.79$\pm5.90$ & 24.75$\pm5.94$ & 25.25$\pm4.81$ & 24.78$\pm10.70$ & 24.95$\pm5.83$ & 21.03$\pm5.43$ & 20.70$\pm4.96$ \\ \bottomrule
    \end{tabular}

    }
    \caption{Full experiment results on GPT-NeoX.}
    \label{tab:full-neox}
\end{table*}

\begin{table*}[ht]
    {\tiny

    \begin{tabular}{lccccccccc}
    \toprule
    \multicolumn{1}{c}{Dataset} & \multicolumn{1}{c}{Metric} & no demo & 0\% & 25\% & 50\% & 75\% & 100\% & random label & shuffled label \\ \midrule
    \multicolumn{1}{l}{\multirow{2}{*}{\begin{tabular}[c]{@{}l@{}}glue\\ sst2\end{tabular}}} & \multicolumn{1}{c}{Accuracy} & 75.46 & 49.40$\pm0.50$ & 61.67$\pm11.16$ & 59.43$\pm7.49$ & 75.83$\pm15.67$ & 90.25$\pm3.86$ & 53.58$\pm4.60$ & 64.04$\pm18.00$ \\
    \multicolumn{1}{l}{} & \multicolumn{1}{c}{F1} & 75.31 & 33.73$\pm1.04$ & 54.18$\pm16.94$ & 51.56$\pm11.75$ & 72.35$\pm22.11$ & 90.20$\pm3.93$ & 41.69$\pm8.58$ & 55.68$\pm25.16$ \\ \hline
    \multicolumn{1}{l}{\multirow{2}{*}{\begin{tabular}[c]{@{}l@{}}glue\\ rte\end{tabular}}} & \multicolumn{1}{c}{Accuracy} & 52.71 & 44.55$\pm5.04$ & 47.15$\pm3.92$ & 48.95$\pm4.12$ & 52.71$\pm3.88$ & 53.72$\pm5.05$ & 51.05$\pm5.71$ & 53.57$\pm3.10$ \\
    \multicolumn{1}{l}{} & \multicolumn{1}{c}{F1} & 34/52 & 38/79$\pm4.09$ & 42.52$\pm6.72$ & 38/34$\pm6.47$ & 48.80$\pm7.75$ & 48.56$\pm8.88$ & 43.63$\pm5.27$ & 48.18$\pm5.01$ \\ \hline
    \multicolumn{1}{l}{\multirow{2}{*}{\begin{tabular}[c]{@{}l@{}}glue\\ mrpc\end{tabular}}} & \multicolumn{1}{c}{Accuracy} & 68.38 & 32.25$\pm2.40$ & 35.98$\pm11.57$ & 43.77$\pm14.81$ & 56.76$\pm14.97$ & 59.71$\pm12.34$ & 43.53$\pm16.03$ & 62.06$\pm6.25$ \\
    \multicolumn{1}{l}{} & \multicolumn{1}{c}{F1} & 40.61 & 27.51$\pm5.58$ & 29.10$\pm7.72$ & 35.84$\pm9.65$ & 44.44$\pm7.97$ & 43.60$\pm2.99$ & 36.11$\pm15.93$ & 43.32$\pm3.11$ \\ \hline
    \multicolumn{1}{l}{\multirow{2}{*}{\begin{tabular}[c]{@{}l@{}}glue\\ wnli\end{tabular}}} & \multicolumn{1}{c}{Accuracy} & 56.34 & 48.45$\pm5.42$ & 47.61$\pm3.51$ & 44.23$\pm5.14$ & 46.20$\pm5.40$ & 46.76$\pm4.92$ & 46.48$\pm6.06$ & 49.58$\pm3.21$ \\
    \multicolumn{1}{l}{} & \multicolumn{1}{c}{F1} & 36.02 & 43.18$\pm7.24$ & 44.84$\pm5.78$ & 38.22$\pm5.67$ & 37.24$\pm7.06$ & 41.39$\pm7.75$ & 38.21$\pm7.87$ & 43.11$\pm6.10$ \\ \hline
    \multicolumn{1}{l}{\multirow{2}{*}{\begin{tabular}[c]{@{}l@{}}super\_glue\\ cb\end{tabular}}} & \multicolumn{1}{c}{Accuracy} & 17.86 & 13.21$\pm5.73$ & 21.07$\pm4.62$ & 40.71$\pm10.37$ & 43.21$\pm12.53$ & 52.86$\pm12.40$ & 20.71$\pm13.87$ & 50.71$\pm8.24$ \\
    \multicolumn{1}{l}{} & \multicolumn{1}{c}{F1} & 15.21 & 10.07$\pm4.34$ & 19.22$\pm4.98$ & 27.78$\pm7.81$ & 27.67$\pm7.59$ & 33.86$\pm11.77$ & 16.36$\pm10.01$ & 27.87$\pm8.82$ \\ \hline
    \multicolumn{1}{l}{\multirow{2}{*}{trec}} & \multicolumn{1}{c}{Accuracy} & 21.60 & 17.92$\pm7.60$ & 30.20$\pm16.30$ & 39.00$\pm15.04$ & 46.88$\pm13.35$ & 49.24$\pm11.47$ & 28.08$\pm5.32$ & 30.44$\pm12.99$ \\
    \multicolumn{1}{l}{} & \multicolumn{1}{c}{F1} & 15.25 & 10.35$\pm3.65$ & 21.42$\pm12.56$ & 26.89$\pm13.70$ & 34.39$\pm10.48$ & 36.45$\pm8.04$ & 18.02$\pm4.39$ & 19.99$\pm9.55$ \\ \hline
    \multicolumn{1}{l}{\multirow{2}{*}{\begin{tabular}[c]{@{}l@{}}financial\\ phrasebank\end{tabular}}} & \multicolumn{1}{c}{Accuracy} & 29.58 & 18.28$\pm4.51$ & 23.31$\pm4.51$ & 23.66$\pm8.88$ & 56.16$\pm12.31$ & 70.95$\pm5.84$ & 20.22$\pm4.78$ & 44.81$\pm18.02$ \\
    \multicolumn{1}{l}{} & \multicolumn{1}{c}{F1} & 34.92 & 17.32$\pm8.45$ & 17.32$\pm8.45$ & 20.07$\pm9.39$ & 41.98$\pm9.71$ & 55.11$\pm12.43$ & 19.06$\pm8.04$ & 27.07$\pm4.36$ \\ \hline
    \multicolumn{1}{l}{\multirow{2}{*}{\begin{tabular}[c]{@{}l@{}}poem\\ sentiment\end{tabular}}} & \multicolumn{1}{c}{Accuracy} & 19.05 & 28.57$\pm20.96$ & 26.67$\pm19.75$ & 42.48$\pm21.26$ & 48.95$\pm19.76$ & 50.86$\pm16.30$ & 34.86$\pm16.74$ & 47.24$\pm21.24$ \\
    \multicolumn{1}{l}{} & \multicolumn{1}{c}{F1} & 19.23 & 17.49$\pm7.04$ & 19.68$\pm8.33$ & 27.39$\pm12.40$ & 26.40$\pm7.73$ & 30.48$\pm7.85$ & 23.50$\pm11.50$ & 30.25$\pm8.53$ \\ \hline
    \multicolumn{1}{l}{\multirow{2}{*}{\begin{tabular}[c]{@{}l@{}}medical\\ questions\_pairs\end{tabular}}} & \multicolumn{1}{c}{Accuracy} & 49.51 & 44.92$\pm4.44$ & 47.18$\pm4.62$ & 50.33$\pm2.90$ & 50.03$\pm1.54$ & 50.92$\pm2.20$ & 50.36$\pm0.99$ & 51.11$\pm1.33$ \\
    \multicolumn{1}{l}{} & \multicolumn{1}{c}{F1} & 33.11 & 36.08$\pm3.17$ & 39.66$\pm6.01$ & 40.11$\pm8.56$ & 38.63$\pm5.83$ & 42.22$\pm8.49$ & 37.51$\pm4.04$ & 38.17$\pm6.85$ \\ \hline
    \multicolumn{1}{l}{\multirow{2}{*}{sick}} & \multicolumn{1}{c}{Accuracy} & 30.51 & 43.80$\pm18.43$ & 50.63$\pm7.97$ & 49.41$\pm9.70$ & 49.45$\pm11.58$ & 57.90$\pm14.12$ & 47.96$\pm13.04$ & 42.79$\pm12.61$ \\
    \multicolumn{1}{l}{} & \multicolumn{1}{c}{F1} & 24.42 & 22.39$\pm6.24$ & 26.71$\pm2.99$ & 27.90$\pm4.82$ & 34.82$\pm12.47$ & 46.76$\pm19.92$ & 26.31$\pm6.83$ & 29.44$\pm6.83$ \\ \hline
    \multicolumn{1}{l}{\multirow{2}{*}{hate\_speech18}} & \multicolumn{1}{c}{Accuracy} & 89.49 & 13.96$\pm6.71$ & 27.09$\pm25.05$ & 46.66$\pm31.79$ & 73.83$\pm15.81$ & 80.48$\pm17.85$ & 63.99$\pm18.21$ & 87.69$\pm1.78$ \\
    \multicolumn{1}{l}{} & \multicolumn{1}{c}{F1} & 47.23 & 12.75$\pm6.00$ & 20.69$\pm14.78$ & 32.54$\pm15.02$ & 45.91$\pm4.15$ & 47.86$\pm4.91$ & 43.98$\pm7.61$ & 47.12$\pm0.36$ \\ \hline
    \multicolumn{1}{l}{\multirow{2}{*}{\begin{tabular}[c]{@{}l@{}}ethos\\ national\_origin\end{tabular}}} & \multicolumn{1}{c}{Accuracy} & 25.29 & 28.05$\pm22.34$ & 35.63$\pm26.32$ & 51.03$\pm19.26$ & 56.09$\pm20.46$ & 68.97$\pm18.90$ & 54.25$\pm25.87$ & 69.89$\pm20.20$ \\
    \multicolumn{1}{l}{} & \multicolumn{1}{c}{F1} & 25.25 & 23.49$\pm16.69$ & 28.11$\pm14.46$ & 40.95$\pm14.75$ & 43.90$\pm10.49$ & 45.34$\pm3.12$ & 41.16$\pm14.06$ & 48.08$\pm9.99$ \\ \hline
    \multicolumn{1}{l}{\multirow{2}{*}{\begin{tabular}[c]{@{}l@{}}ethos\\ race\end{tabular}}} & \multicolumn{1}{c}{Accuracy} & 32.18 & 22.07$\pm0.51$ & 43.68$\pm26.06$ & 48.05$\pm19.22$ & 65.75$\pm9.56$ & 78.16$\pm0.00$ & 55.17$\pm21.12$ & 78.16$\pm0.00$ \\
    \multicolumn{1}{l}{} & \multicolumn{1}{c}{F1} & 31.86 & 18.25$\pm0.72$ & 34.13$\pm17.43$ & 41.94$\pm14.12$ & 49.77$\pm5.11$ & 43.87$\pm0.00$ & 43.53$\pm11.74$ & 43.87$\pm0.00$ \\ \hline
    \multicolumn{1}{l}{\multirow{2}{*}{\begin{tabular}[c]{@{}l@{}}ethos\\ religion\end{tabular}}} & \multicolumn{1}{c}{Accuracy} & 29.89 & 19.54$\pm1.63$ & 28.97$\pm10.76$ & 30.57$\pm13.74$ & 69.43$\pm13.35$ & 80.00$\pm2.52$ & 51.05$\pm24.01$ & 77.01$\pm3.90$ \\
    \multicolumn{1}{l}{} & \multicolumn{1}{c}{F1} & 29.74 & 17.18$\pm1.30$ & 26.89$\pm11.02$ & 28.79$\pm13.64$ & 46.93$\pm3.16$ & 46.52$\pm5.17$ & 37.60$\pm11.80$ & 47.83$\pm3.62$ \\ \hline
    \multicolumn{1}{l}{\multirow{2}{*}{\begin{tabular}[c]{@{}l@{}}tweet\_eval\\ hate\end{tabular}}} & \multicolumn{1}{c}{Accuracy} & 42.70 & 44.02$\pm6.56$ & 47.76$\pm5.18$ & 53.08$\pm5.71$ & 55.76$\pm4.35$ & 59.72$\pm2.77$ & 54.74$\pm2.32$ & 54.42$\pm4.48$ \\
    \multicolumn{1}{l}{} & \multicolumn{1}{c}{F1} & 29.92 & 40.54$\pm4.18$ & 43.40$\pm3.39$ & 42.43$\pm4.22$ & 48.60$\pm8.50$ & 49.67$\pm9.04$ & 42.19$\pm7.08$ & 46.74$\pm6.60$ \\ \hline
    \multicolumn{1}{l}{\multirow{2}{*}{\begin{tabular}[c]{@{}l@{}}tweet\_eval\\ stance\_atheism\end{tabular}}} & \multicolumn{1}{c}{Accuracy} & 25.00 & 20.00$\pm2.58$ & 21.92$\pm1.05$ & 22.31$\pm2.19$ & 28.55$\pm10.88$ & 45.38$\pm17.50$ & 20.00$\pm2.58$ & 43.85$\pm16.06$ \\
    \multicolumn{1}{l}{} & \multicolumn{1}{c}{F1} & 17.82 & 15.57$\pm5.70$ & 13.37$\pm2.60$ & 14.86$\pm2.85$ & 20.43$\pm11.06$ & 29.66$\pm11.31$ & 13.41$\pm2.69$ & 25.86$\pm6.71$ \\ \hline
    \multicolumn{1}{l}{\multirow{2}{*}{\begin{tabular}[c]{@{}l@{}}tweet\_eval\\ feminist\end{tabular}}} & \multicolumn{1}{c}{Accuracy} & 49.51 & 44.92$\pm4.44$ & 47.18$\pm4.62$ & 50.33$\pm2.90$ & 50.03$\pm1.54$ & 50.92$\pm2.20$ & 50.36$\pm0.99$ & 51.11$\pm1.33$ \\
    \multicolumn{1}{l}{} & \multicolumn{1}{c}{F1} & 33.11 & 36.08$\pm3.17$ & 39.66$\pm6.01$ & 40.11$\pm8.56$ & 38.63$\pm5.83$ & 42.22$\pm8.49$ & 37.51$\pm4.04$ & 38.17$\pm6.85$ \\ \bottomrule
    \end{tabular}
    }
    \caption{Full experiment results on GPT-J.}
    \label{tab:full-j}
\end{table*}

\end{document}